\documentclass[12pt]{iopart}

\usepackage{iopams}  
\usepackage[hidelinks]{hyperref} %
\usepackage{graphicx} 
\usepackage{setstack} 
\usepackage{fontenc}
\usepackage[T1]{fontenc}
\usepackage{subcaption}
\usepackage{booktabs}

\usepackage[style=numeric,sorting=none,giveninits=true]{biblatex}
\AtEveryBibitem{%
  \DeclareFieldFormat[article]{title}{#1}%
  \DeclareFieldFormat[inproceedings]{title}{#1}%
  \DeclareFieldFormat[incollection]{title}{#1}%
  \DeclareFieldFormat[inbook]{title}{#1}%
  \DeclareFieldFormat[book]{title}{#1}%
}

\renewbibmacro{in:}{}

\DeclareUnicodeCharacter{03BC}{\textemdash}

\begin{document}

\title[{Real-Time Neuromorphic Navigation}]{Real-Time Neuromorphic Navigation: Guiding Physical Robots with Event-Based Sensing and Task-Specific Reconfigurable Autonomy Stack}

\author{
    Sourav Sanyal\textsuperscript{†}, Amogh Joshi\textsuperscript{†}, Adarsh Kosta,  Kaushik Roy\\
    Electrical and Computer Engineering, Purdue University, USA
}

\ead{kaushik@purdue.edu}
\footnotetext{These authors contributed equally.}


\begin{abstract}
Neuromorphic vision, inspired by biological neural systems, has recently gained significant attention for its potential in enhancing robotic autonomy. This paper presents a systematic exploration of a proposed Neuromorphic Navigation framework that uses event-based neuromorphic vision to enable efficient, real-time navigation in robotic systems. We discuss the core concepts of neuromorphic vision and  navigation, highlighting their impact on improving robotic perception and decision-making. The proposed reconfigurable Neuromorphic Navigation framework adapts to the specific needs of both ground robots (Turtlebot) and aerial robots (Bebop2 quadrotor), addressing the task-specific design requirements (algorithms) for optimal performance across the autonomous navigation stack -- Perception, Planning, and Control. We demonstrate the versatility and the effectiveness of the framework through two case studies: a Turtlebot performing local replanning for real-time navigation and a Bebop2 quadrotor navigating through moving gates. Our work provides a scalable approach to task-specific, real-time robot autonomy leveraging neuromorphic systems, paving the way for energy-efficient autonomous navigation.
\end{abstract}

%
%
%
%
%

\section{Introduction}
Robotic navigation in dynamic environments poses significant challenges due to high computational demands, latency, and power consumption associated with conventional frame-based sensors and deep learning-based perception systems. These systems process dense, redundant data and often struggle to provide reliable performance under varying lighting and motion conditions. This makes real-time, energy-efficient navigation in visually challenging environments an important problem for resource-constrained robotic platforms. Neuromorphic navigation, inspired by biological neural systems, offers a promising solution by leveraging event-driven sensing and spiking neural networks (SNNs) \cite{Roy2019} to complement existing approaches, potentially increasing energy efficiency through reduced computation, making them well-suited for real-time robotic tasks, particularly when prompt reaction and responsive adaptation is desirable.
Neuromorphic computing, inspired by the architecture and dynamics of biological neural systems, represents a fundamental shift in how computational platforms perceive, process, and respond to the world \cite{Schuman2022, Furber2016}. By employing event-driven, asynchronous, and energy-efficient computation, neuromorphic systems are well-suited to latency-sensitive and resource-constrained robotics applications, making them particularly attractive for real-time navigation tasks.

In contrast to traditional frame-based vision systems—which frequently suffer from high latency, motion blur, excessive power consumption, and limited temporal resolution—\emph{neuromorphic vision} has emerged as an alternative paradigm. Built on neuromorphic principles, event-based vision provides a more efficient and responsive approach to visual perception in dynamic environments. Rather than capturing entire frames at fixed intervals, these systems detect only changes in scene intensity. This approach reduces data redundancy, lowers power requirements, and delivers the rapid, continuous environmental feedback essential for high-speed obstacle avoidance and adaptive trajectory replanning. At the core of neuromorphic vision lie \emph{Dynamic Vision Sensors (DVS)}, also known as \emph{Event Cameras}, which record only intensity changes rather than full image frames \cite{dvs1,dvs2,dvs3}. Operating at microsecond-level temporal resolution, DVS devices substantially cut down on unnecessary data transmission and bandwidth usage, resulting in faster and more efficient perception.

However, integrating event-based vision into robotic perception poses challenges. Conventional computer vision algorithms, optimized for frame-based inputs, often fail to exploit the rich temporal information in event streams. Spiking Neural Networks (SNNs) \cite{Roy2019} address this by processing asynchronous, time-varying signals as discrete spikes, thereby handling DVS outputs efficiently \cite{lee2020spike, kosta2023adaptive}. Their parallel, event-driven computation is especially useful for tasks such as optical flow estimation, segmentation, and object detection \cite{lee2020spike, kosta2023adaptive, halsie, nagaraj2023dotie}, enabling rapid feature extraction and real-time, energy-efficient decision-making \cite{evplanner, joshi2024, Lichtsteiner2008}.

In addition, several works adopt a hybrid approach by merging spiking layers with conventional ANN modules. For example, \cite{bobw, fusionflow} integrate event-driven and frame-based pipelines for optical flow estimation, while \cite{evplanner, joshi2024, sanyal2025} combine spiking object detection with physics-based neural networks for energy-efficient velocity prediction in trajectory planning. Further, \cite{joshi2025neuro} takes the neuromorphic paradigm a step further by integrating an interactive front-end language model with spiking-based perception, enabling human-in-the-loop navigation and decision-making. These developments demonstrate that neuromorphic computing is rapidly evolving beyond traditional SNNs to embrace diverse, hybrid architectures.

Building on these principles, this paper introduces the concept of \emph{Neuromorphic Navigation}, integrating neuromorphic vision—characterized by event-driven sensing and processing—with a generic yet task-customizable  autonomous navigation stack. By combining neuromorphic vision-based perception with physics-informed planning and control techniques, we enable robust performance in real-world robotic applications. Drawing from the principles of event-driven computation and leveraging the joint efficiency of SNNs and DVS technology, \emph{Neuromorphic Navigation} harnesses the underlying physics of robot kinematics to deliver stable, agile, and energy-efficient navigation across diverse robotic platforms. 
Robotic tasks vary widely: ground robots require precise, deliberate motion in cluttered terrains \cite{gsnn} while aerial platforms demand rapid, high-speed agility in dynamic settings \cite{evplanner, joshi2024}. Ground platforms, such as the Turtlebot, can exploit adaptive event-driven perception to achieve precise, real-time local replanning, whereas aerial robots can benefit from microsecond-scale reaction times and low-power processing for swift maneuvering and stable operation. Addressing such diverse requirements necessitates a flexible framework that reuses core modules while allowing for reconfiguration to meet each application’s demands.
Our proposed \emph{Reconfigurable Neuromorphic Navigation Framework} achieves this balance through a modular architecture. 
 Our key contributions are:

\begin{itemize}
    \item \textbf{Neuromorphic Perception for Real-Time Navigation:} We leverage event-based vision sensors and spiking neural networks (SNNs) to enable low-latency, energy-efficient perception and decision-making for robotic navigation.
    
    \item \textbf{Reconfigurable Neuromorphic Navigation Stack:} We propose a modular autonomy framework that  integrates perception, physics-informed planning, and control, adaptable across different robotic platforms,  including ground (TurtleBot) and aerial (Bebop2 quadrotor) robots.
    
    \item \textbf{Physics-Informed Planning for Robust and Energy-Efficient Control:} We incorporate physics-guided neural networks (PgNNs) and physics-informed motion constraints to optimize trajectories, ensuring dynamically feasible and energy-efficient navigation.
    
    \item \textbf{Experimental Validation in Dynamic Environments:} We conduct both simulations and real-world experiments, demonstrating neuromorphic navigation in challenging, dynamic scenarios, such as obstalce avoidance  by ground robot and  drone navigation through a moving ring-shaped gate.
    
    \item \textbf{Edge Deployment:} We implement our approach on an NVIDIA Jetson Nano, showcasing real-time deployment for resource-constrained robotics applications.
\end{itemize}

This work highlights the advantages of neuromorphic computing for robotic autonomy, providing a scalable and efficient solution for real-time, adaptive navigation in diverse and dynamic environments. The remainder of this paper is organized as follows: Section \ref{background} introduces the foundations of neuromorphic computing, event-based vision, and autonomous navigation. In Section \ref{neuromorphic_stack}, we present the reconfigurable neuromorphic navigation autonomy stack, detailing how its perception, planning, and control layers can be adapted for diverse operational requirements. Section \ref{taxonomical} illustrates the effectiveness and versatility of the framework in simulated environments as well as real-world scenarios. Finally in Section \ref{conclusion} we conclude the paper with insights into the future of neuromorphic navigation.

\section{Background and Preliminaries}\label{background}
\subsection{What is Neuromorphic Computing?}

Neuromorphic computing takes inspiration from the organizational principles of biological nervous systems to enable efficient, low-power, and event-driven information processing. Instead of relying on the continuous and typically frame-based computations of conventional Artificial Neural Networks (ANNs), neuromorphic systems leverage time-encoded signals and local event-driven computations to mimic key features of biological neural circuits.
In conventional ANNs, neurons generally compute a weighted sum of their inputs and then apply a continuous activation function. For instance, consider a single artificial neuron:
\begin{equation}
y = f\left(\sum_{i} w_i x_i + b\right),
\end{equation}
where \(x_i\) are the input signals, \(w_i\) are the corresponding weights, \(b\) is a bias term, and \(f(\cdot)\) is a nonlinear activation function (e.g., sigmoid, ReLU). Such networks typically process information in discrete time steps (frames), passing entire batches of data through all neurons synchronously. Although conventional ANNs are inspired by the brain's structure, they do not capture the asynchronous, event-driven behavior observed in biological systems. Neuromorphic computing, on the other hand, is designed to process sparse, temporally precise spike events, which not only reduces redundant computations but also enables low-latency and energy-efficient processing suitable for real-time applications.

\begin{figure}[t]
    \centering
    \includegraphics[width=0.9\textwidth]{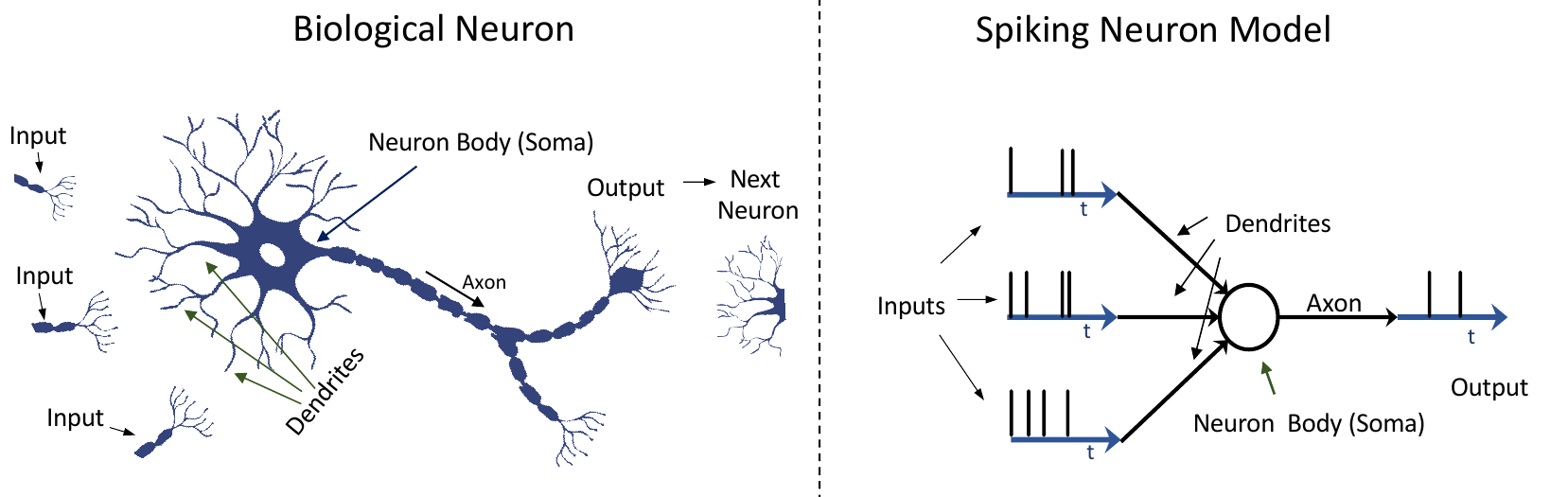}
    \caption{Comparison between biological neurons and spiking neuron models, highlighting dendrites, soma, axons, and the discrete spike timing (t) in Spiking Neural Networks (SNNs).}
    \label{fig:biological_inspiration}
\end{figure}

Figure \ref{fig:biological_inspiration} shows how neuromorphic computing draws directly from biological neurons. In biological systems, information is communicated via electrical impulses (spikes) that occur at specific points in time, rather than through continuous-valued signals. This timing-based communication is at the heart of \emph{Spiking Neural Networks (SNNs)}, the key kernel of neuromorphic computing.

SNNs replace continuous activation values with discrete spike events. A popular model for a spiking neuron is the Leaky Integrate-and-Fire (LIF) neuron, which maintains a membrane potential \(V(t)\) that evolves over time:
\begin{equation}
\tau_m \frac{dV}{dt} = -V(t) + R I(t),
\end{equation}
where \(\tau_m\) is the membrane time constant, \(R\) is the input resistance, and \(I(t)\) is the synaptic input current resulting from incoming spikes. When \(V(t)\) crosses a threshold \(V_{{th}}\), the neuron emits a spike and resets its membrane potential to a baseline value. In this way, SNN neurons operate asynchronously and exploit the exact timing of incoming and outgoing spikes:
\begin{equation}
s_i(t) = \sum_{k} \delta(t - t_i^k),
\end{equation}
where \( t_i^k \) are the spike times of neuron \( i \) and \(\delta(\cdot)\) is the Dirac delta function. This representation allows SNNs to naturally encode and process temporal information, making them highly efficient for tasks involving continuous streams of data.

\begin{figure}[t]
    \centering
    \includegraphics[width=0.8\textwidth]{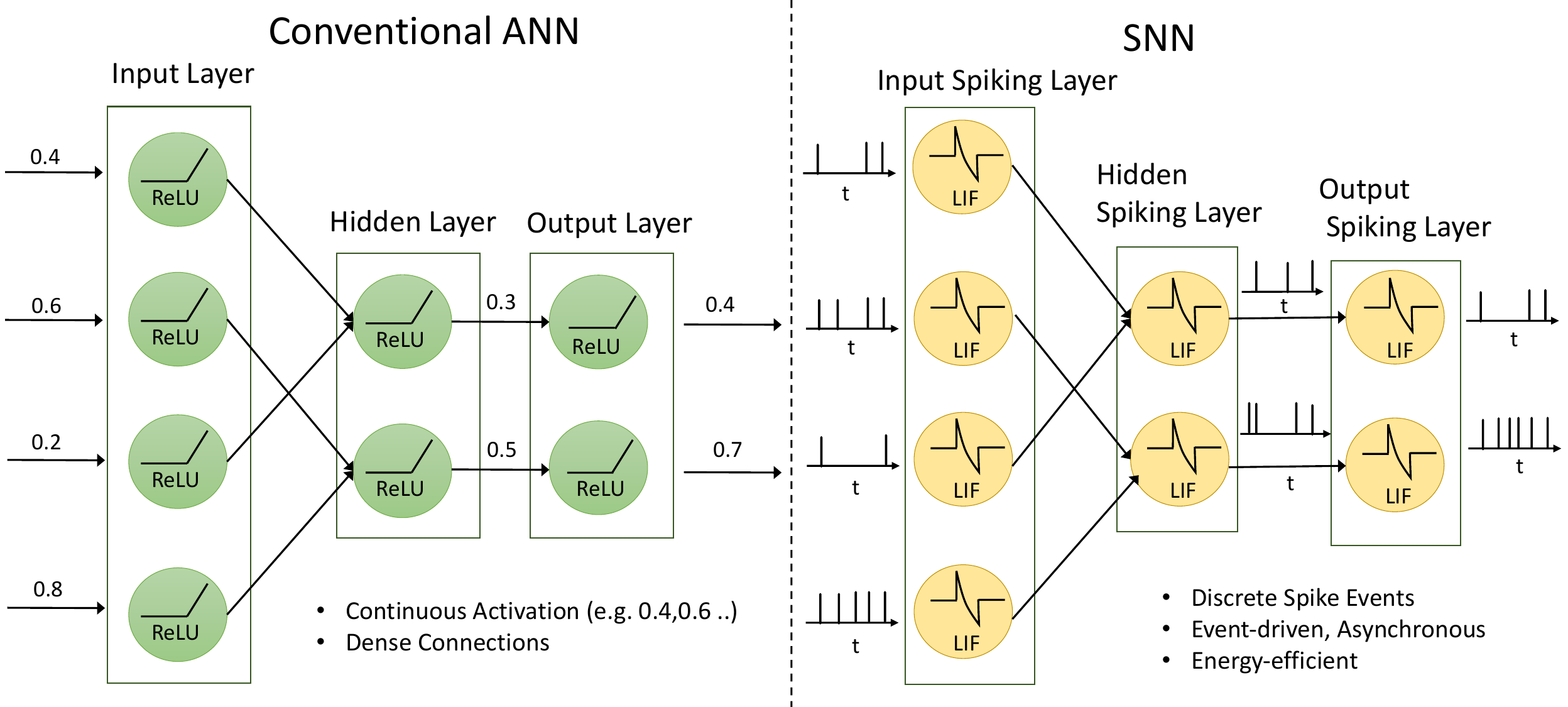}
    \caption{Side-by-side comparison of conventional Artificial Neural Networks (ANNs) and Spiking Neural Networks (SNNs), showcasing continuous activation in ANNs versus discrete, event-driven spikes in SNNs.}
    \label{fig:ann_snn_comparison}
\end{figure}

Figure \ref{fig:ann_snn_comparison} highlights the contrast between ANNs and SNNs. While ANNs rely on continuous signals and synchronous data processing, SNNs employ discrete spikes and event-driven updates. Such architectures are inherently low-power because computation and communication occur only when meaningful events (spikes) are present. This property aligns perfectly with hardware implementations like Intel’s Loihi \cite{Davies2018} and IBM’s TrueNorth \cite{Merolla2014}, which facilitate  SNN deployments in asynchronous data settings \cite{Indiveri2015, Lee2016}.

Neuromorphic computing has shown remarkable potential in robotics and other domains demanding real-time, adaptive responses. SNNs have demonstrated success in optical flow estimation \cite{Benosman2014, Bardow2016}, trajectory planning \cite{kosta2023adaptive}, and real-time control \cite{halaly2024continuous}. Moreover, advanced training methods inspired by biological learning rules, such as Spike-Timing-Dependent Plasticity (STDP) and surrogate gradient approaches \cite{Tavanaei2019, Pfeiffer2018}, enable SNNs to learn efficiently in dynamic and unstructured environments \cite{Gallego2022, zhang2023dynamic}. These innovations have further expanded the reach of neuromorphic computing to tasks like Imitation Learning and Evolutionary Learning \cite{Schuman_2022}, which contrast the strengths of different learning paradigms in neuromorphic control systems.
Furthermore, studies such as \cite{Schuman_2022} highlight the exploration of evolutionary and imitation learning paradigms for neuromorphic systems, underscoring the adaptability and robustness of SNNs in addressing edge-computing challenges. This adaptability makes neuromorphic computing an essential platform for next-generation robotics and autonomous systems.

\begin{figure}[t]
    \centering
    \includegraphics[width=1.0\textwidth]{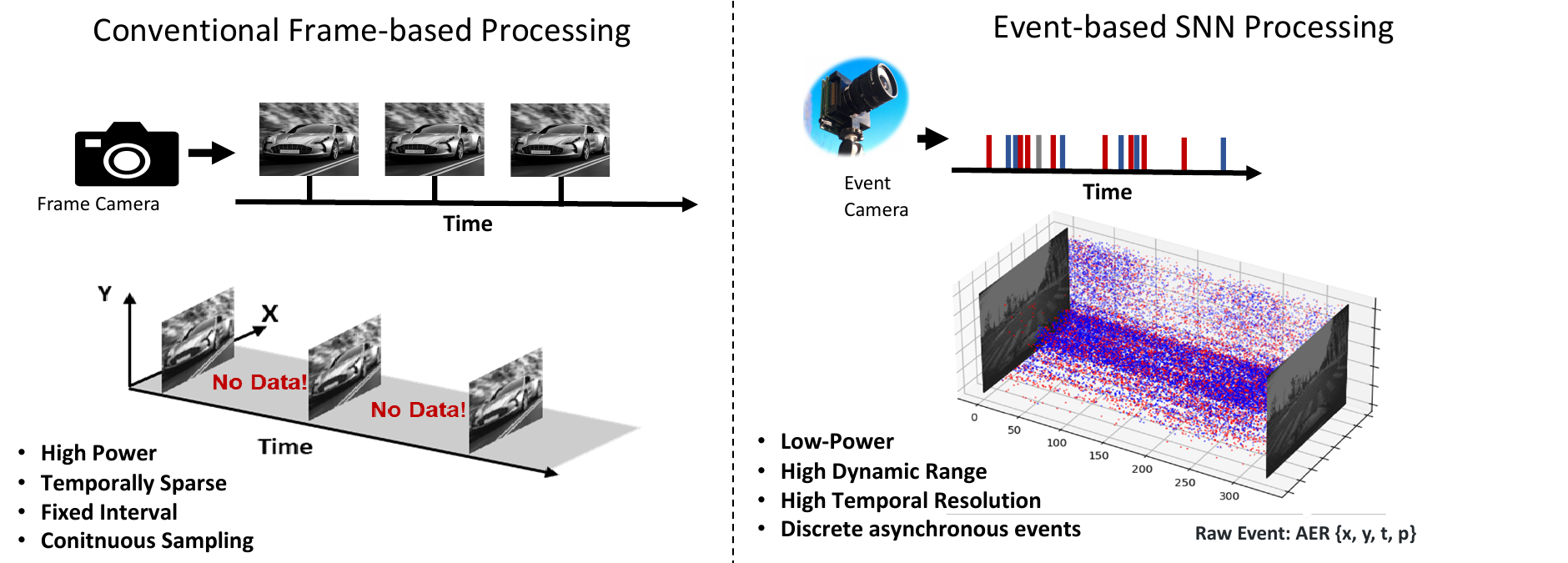}
    \caption{Comparison of conventional frame-based vision processing and event-based SNN processing, highlighting the low-power, high-temporal-resolution advantages of Dynamic Vision Sensors (DVS).}
    \label{fig:frame_vs_event_processing}
\end{figure}

\subsection{Neuromorphic Perception in Robotics}

Event-based vision offers a fundamentally different paradigm from conventional frame-based imaging. Instead of capturing entire frames at discrete intervals, \emph{Dynamic Vision Sensors (DVS)}—also known as event cameras—detect and record pixel-level changes in brightness asynchronously. Each such change generates an \emph{event}:
\begin{equation}
e_k = (x_k, y_k, t_k, p_k),
\end{equation}
where \((x_k, y_k)\) denotes the event’s pixel coordinates, \(t_k\) its precise time of occurrence, and \(p_k \in \{+1, -1\}\) the polarity of the intensity change (increase or decrease in brightness). This event-driven representation inherently reduces redundancy, lowers power consumption, and enables ultra-fast response times, making DVS ideal for scenarios with high dynamic range and variable lighting conditions \cite{Lichtsteiner2008, Posch2011, Brandli2014}.

Figure \ref{fig:frame_vs_event_processing} contrasts frame-based vision, which captures entire scenes at fixed intervals, with the event-based approach, where data are produced only when changes occur. Unlike frame-based methods that can waste resources on unchanging regions, event-driven systems naturally focus on salient, time-varying features. Integrated with Spiking Neural Networks (SNNs), which process information as discrete spike trains, this event-based paradigm enables high temporal resolution and energy efficiency. Such capabilities have proven indispensable in real-time robotics tasks, including visual odometry \cite{Zhu2017}, SLAM \cite{rebecq2018emvs}, and obstacle avoidance \cite{Falanga2020}, where timely, low-latency perception is critical.

The synergy between DVS and SNNs also extends to more advanced vision tasks. For instance, constructing dense optical flow or stereo depth maps from sparse, asynchronous events may initially seem challenging. Yet, by aggregating events over short time intervals, it is possible to estimate local motion fields \cite{EVFlowNet2018, ParedesValles2019, Zhu2019}. This often involves mapping the spatiotemporal distribution of events onto SNN layers, where spike timings encode motion cues. Furthermore, real-time 3D reconstruction and six degrees-of-freedom (6-DoF) tracking with event cameras exemplify the potential of event-based vision in dynamic environments \cite{Kim2016}.  Additionally, tasks like neuromorphic stereo depth estimation \cite{risi2021instantaneous}, depth estimation on dedicated neuromorphic hardware \cite{chiavazza2023low}, and dynamic object detection \cite{nagaraj2023dotie, IMO2023} demonstrate the versatility of event-based processing. These methods allow autonomous systems to achieve low-latency perception with minimal computational overhead, paving the way for sustainable and robust vision-driven applications.

Neuromorphic systems have demonstrated significant success in tasks such as \emph{optical flow estimation}, where motion patterns are extracted directly from event streams, bypassing redundant computations \cite{lee2020spike, fusionflow, kosta2023adaptive, bobw}. Similarly, event-based methods for \emph{unsupervised depth estimation} and \emph{egomotion estimation} utilize the high temporal resolution of DVS outputs to generate real-time depth maps and motion trajectories \cite{zhu2019unsupervised, ye2020unsupervised}. Event-based SLAM frameworks \cite{rideg2023event, yu2019neuroslam, zeng2020neurobayesslam} and visual odometry solutions \cite{zhou2021event, chen2023esvio} highlight the robustness of neuromorphic perception for localization and mapping under resource-constrained conditions.
Techniques such as \emph{contrast maximization} \cite{gallego2018unifying} and reward-based refinements \cite{stoffregen2019event} have further improved feature extraction and motion estimation, showcasing the flexibility of neuromorphic vision systems. The emerging hybrid frameworks, such as Adaptive-SpikeNet \cite{kosta2023adaptive}, combine SNN layers with traditional ANN models to optimize performance and efficiency.
\emph{Object detection} is another key application of neuromorphic perception. By directly processing event streams, neuromorphic systems focus on regions of significant motion and filter out irrelevant static background data, ensuring high responsiveness and energy efficiency \cite{nagaraj2023dotie}. In this work, we employ object detection as the core neuromorphic perception method to develop a modular neuromorphic navigation stack for real-world robotic agents.

\begin{figure}[t]
    \centering
    \includegraphics[width=0.99\textwidth]{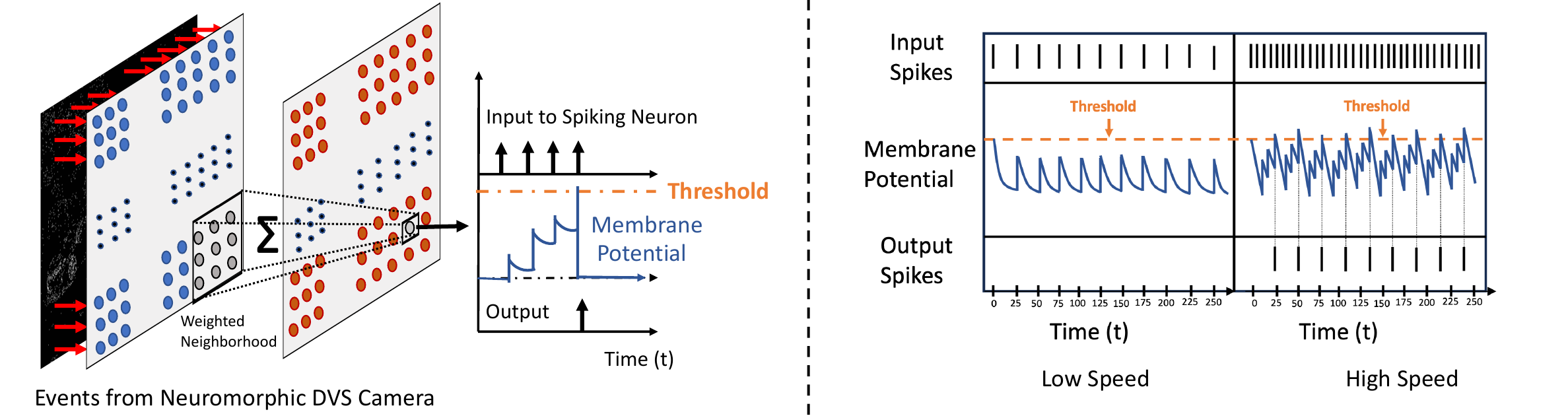}
    \caption{\textbf{left:} Events are grouped spatially using a weighted neighborhood, and spiking neurons generate outputs for object detection and tracking. \textbf{right:} Input spikes are temporally filtered by spiking neurons; only dense event streams exceed the threshold, producing output spikes.}
    \label{fig:object_detection_architecture}
    \label{fig:snn_object_detection}
\end{figure}

Figure \ref{fig:object_detection_architecture} illustrates this process. Temporal filtering using spiking neurons (Figure \ref{fig:snn_object_detection}) accumulates input spikes from dynamic events. When a sufficient number of spikes occur, the membrane potential crosses a threshold, triggering output spikes. Subsequently, events are spatially grouped using weighted clustering techniques (Figure \ref{fig:snn_object_detection}), allowing SNNs to generate outputs for object detection and tracking efficiently.
\begin{figure*}[t]
    \centering
    \includegraphics[width=0.95\textwidth]{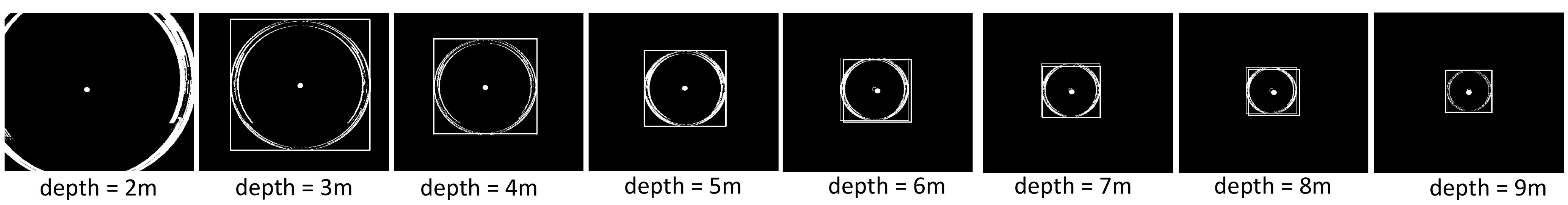}
    \caption{%
        \textbf{Event-based Object Detection at Various Depths:} 
        Each sub-panel shows neuromorphic event output and a bounding box 
        around a moving gate as it recedes from the camera (left to right). 
        Although the event density decreases with increasing depth, 
        the SNN isolates and tracks the gate in real time.
    }
    \label{fig:ev_box}
\end{figure*}

\subsection{Autonomous Navigation: Planning and Control}

Autonomous navigation fundamentally relies on three interconnected components: \emph{perception}, \emph{planning}, and \emph{control}. Building on the discussed foundations of neuromorphic perception—enabling event-driven, bio-inspired processing of sensory data \cite{lee2020spike, Gallego2022, yang2023neuromorphic, novo2024neuromorphic}—recent efforts have extended neuromorphic principles into planning and control. These developments aim to improve trajectory optimization, enhance energy efficiency, and achieve real-time adaptability in complex, dynamic environments.

\subsubsection{Neuromorphic Planning:}

Robotic planning involves computing trajectories that balance constraints such as obstacle avoidance, speed, and energy consumption. Traditional path-planning algorithms, including Rapidly-exploring Random Trees (RRT) \cite{RRTStar2022}, have been integrated into neuromorphic frameworks, facilitating more efficient, scalable solutions and enabling applications like Simultaneous Localization and Mapping (SLAM) in event-driven contexts \cite{vSLAM2018, tang2019spiking}. These neuromorphic adaptations align well with asynchronous data streams and temporal sparsity, allowing robots to handle dynamic changes in their environments more effectively.To further improve efficiency, recent physics-informed models incorporate robot dynamics and environmental constraints directly into the planning process, improving trajectory optimization for both underactuated and overactuated robotic systems \cite{rampnet, bianchi2024physics, chee2022knode}. These physics-informed approaches allow for energy-saving maneuvers and rapid adjustments to evolving conditions.
Neuromorphic planning has also drawn inspiration from biological spatial representations observed in the hippocampus and entorhinal cortex. Spatial encoding via place cells and grid cells \cite{Moser2008, tang2017cognitive} supports cognitive mapping and memory-based navigation strategies, providing robots with robust methods of topological navigation and long-term adaptability. Additionally, hybrid computational frameworks combining spiking neural network (SNN) layers with conventional artificial neural networks \cite{kosta2023adaptive} have demonstrated the potential to improve both accuracy and efficiency, paving the way for scalable neuromorphic navigation across diverse scenarios.

\subsubsection{Neuromorphic Control:}

Once trajectories are established, control systems must convert planned paths into precise actuator commands. Neuromorphic control leverages SNNs and bio-inspired learning rules to achieve adaptive, low-latency responses under diverse conditions \cite{Neftci2019, brunner2020trajectory}. Early work focused on robotic manipulators \cite{helferty1991neuromorphic} and arms \cite{polykretis2023bioinspired}, with later studies extending to bipedal robots \cite{folgheraiter2019neuromorphic}. These advances have been enhanced by neuromorphic PID controllers \cite{stagsted2023towards}, closed-loop spiking systems \cite{zhao2020closed}, and mixed-signal neuromorphic processors \cite{ma2017darwin}. Adaptive components using STDP \cite{zhao2015spike} and on-chip hardware adaptations \cite{glatz2019adaptive, yang2018real} further illustrate the scalability of these architectures, with recent work highlighting SNN-based control across various robotic tasks \cite{mitchell2017neon, szczecinski2023perspective, aitsam2022neuromorphic, blum2017neuromorphic}. A notable distinction is between general SNN-driven control and DVS-based SNN control; the latter employs event-driven visual input to guide rapid decision-making \cite{cheng2020neuromorphic, sandamirskaya2022neuromorphic, zhao2020neuromorphic}, enabling instantaneous adaptation to visual stimuli.

\subsubsection{Embodied Neuromorphic Intelligence:}

Integrating neuromorphic perception, planning, and control within physical platforms is central to the concept of \emph{Embodied Neuromorphic Intelligence} \cite{bartolozzi2022embodied}. This approach embeds neuromorphic processors and closed-loop spiking controllers \cite{zhao2020closed} into real-world robots, enabling energy-efficient, responsive, and scalable solutions. Such embodied systems can exploit temporal sparsity, asynchronous operation, and bio-inspired learning, resulting in robots capable of seamlessly interacting with dynamic environments, optimizing computation, and reducing power consumption.
By harnessing the synergy between neuromorphic perception, physics-informed planning, and spike-driven control, autonomous robotic systems stand to gain unprecedented adaptability, efficiency, and robustness. This integrated neuromorphic paradigm promises to redefine the capabilities of autonomous navigation and to inspire continued innovation in real-time, bio-inspired robotic intelligence.

While neuromorphic navigation frameworks show great promise, several technical challenges remain. Current algorithms for event-based sensors and neuromorphic processing are still evolving and must better exploit the temporal sparsity of event-driven data \cite{Gallego2022}. Efficient training methods for spiking neural networks (SNNs) are needed to bridge the gap with traditional deep learning \cite{Tavanaei2019}. Additionally, integrating neuromorphic components with existing robotic platforms poses challenges in sensor fusion and actuator control \cite{sandamirskaya2022neuromorphic}. Continued research in these areas is essential to fully realize the potential of neuromorphic navigation in dynamic environments.

\begin{table*}[t]
\footnotesize
    \centering
    \caption{Comparison with Existing Neuromorphic Navigation Works}
    \label{tab:comparison}
    \renewcommand{\arraystretch}{1.1}
    \begin{tabular}{@{}p{4.0cm}p{5cm}p{5.7cm}@{}}
        \toprule
        \textbf{Prior Works} & \textbf{Limitations} & \textbf{This Work} \\
        \midrule
        Perception \cite{Gallego2022, Lichtsteiner2008}, Planning \cite{yu2019neuroslam, rideg2023event}, Control \cite{sandamirskaya2022neuromorphic}  
        & Studied in isolation; lacks a unified approach integrating perception, planning, and control. 
        & Full-stack solution combining perception, planning, and control into a task-specific, adaptable framework. \\
        \midrule
        Heuristic-based planning \cite{yu2019neuroslam, rideg2023event} 
        & Relies on hand-crafted rules; struggles with dynamic environments and energy optimization. 
        & Physics-informed trajectory planning ensures dynamically feasible and energy-efficient navigation. \\
        \midrule
        Platform-specific (Quadrotors \cite{rampnet}, Ground Robots \cite{EVFlowNet2018})  
        & Limited to specific robotic platforms; lacks generalizability. 
        & Scalable across both aerial and ground robots using a reconfigurable perception-planning-control stack. \\
        \bottomrule
    \end{tabular}
\end{table*}

\subsection{Comparison with Existing Work}
Neuromorphic navigation has been explored extensively, particularly in event-based perception and planning. However, most existing methods address isolated aspects, such as neuromorphic sensing \cite{Gallego2022, Lichtsteiner2008}, SNN-based perception \cite{kosta2023adaptive, lee2020spike}, or event-driven control \cite{stagsted2023towards, sandamirskaya2022neuromorphic}. In contrast, our work presents a holistic, task-specific neuromorphic autonomy stack that integrates perception, physics-informed planning, and real-time adaptive control for efficient and scalable robot navigation. Below, we outline the key differences between our approach and prior methods:

\begin{itemize}
    \item \textbf{End-to-End Neuromorphic Navigation Stack:} While prior works have studied \emph{event-based perception} \cite{nagaraj2023dotie, Bardow2016} and \emph{spiking neural network (SNN) processing} \cite{kosta2023adaptive}, these are often treated as standalone components. Our approach combines event-based sensing with neuromorphic planning and control into a unified system, ensuring seamless interaction across perception, decision-making, and actuation.

    \item \textbf{Physics-Informed Planning for Dynamic Adaptability:} Our work incorporates physics-based constraints into trajectory planning, ensuring dynamically feasible and energy-efficient navigation, unlike previous neuromorphic SLAM \cite{yu2019neuroslam, rideg2023event} and planning methods that rely solely on heuristic-based approaches.

    \item \textbf{Scalable Across Multiple Robotic Platforms:} Our work demonstrates adaptability across both aerial and ground robots, using the same perception-planning-control pipeline, enabling \emph{a task-specific, reconfigurable autonomy stack}.
\end{itemize}

Finally, while neuromorphic algorithms have been validated in simulation or on cloud-based processing units substantially, real-time deployment on \emph{resource-constrained hardware} remains limited. We demonstrate real-time execution on an NVIDIA Jetson Nano, showcasing practical feasibility for \emph{low-power embedded robotic platforms}.

\section{Neuromorphic Navigation Autonomy Stack}
\label{neuromorphic_stack}
\begin{figure}[htbp]
    \centering
    \includegraphics[width=0.9\linewidth]{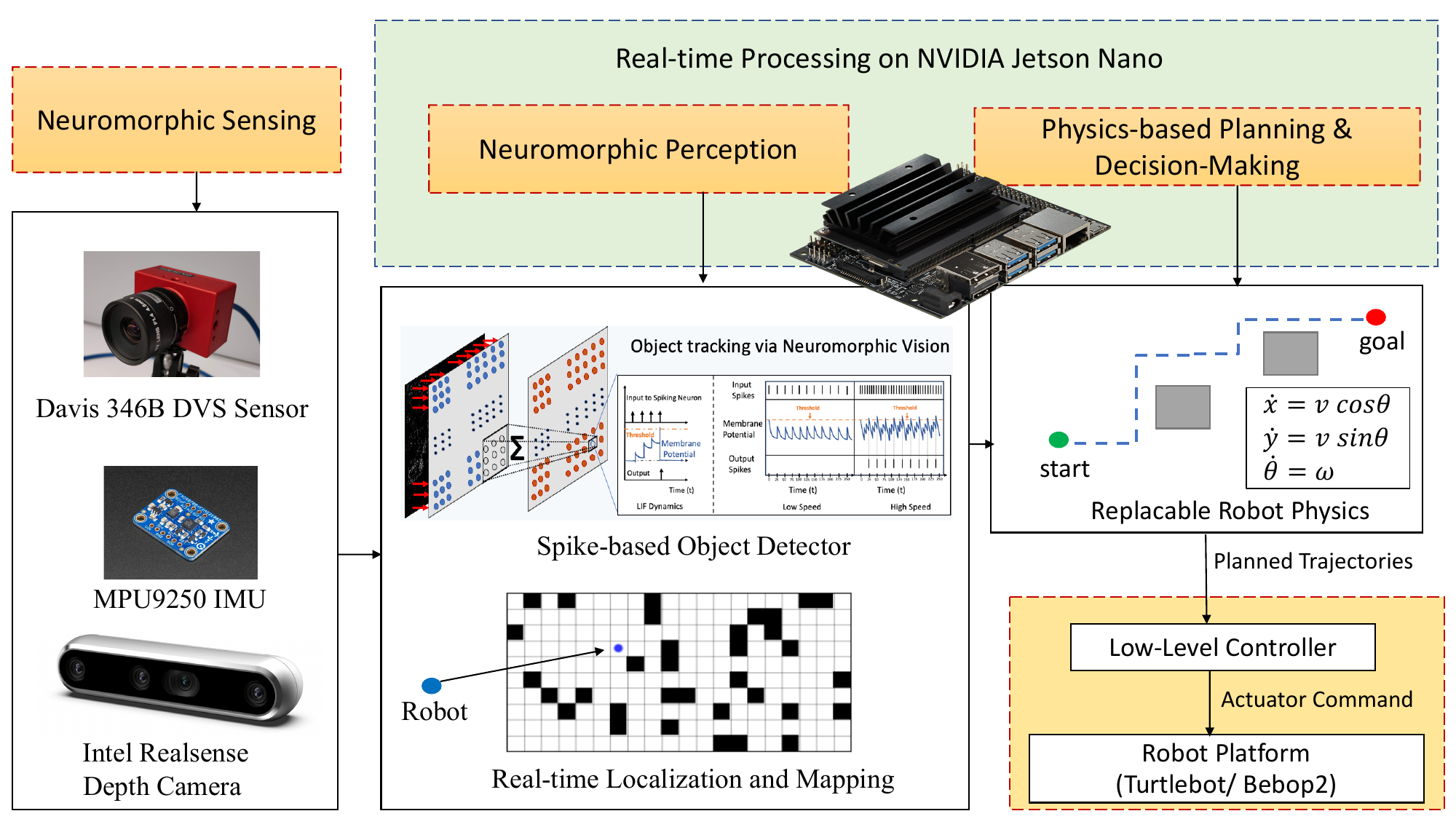}
    \caption{\textbf{Neuromorphic navigation framework}: Neuromorphic sensors 
    (DVS camera, IMU, and depth camera) provide input to spike-based perception 
    modules for object detection and real-time localization/mapping. Planning 
    and decision-making, including the robot’s motion physics, occur on the 
    NVIDIA Jetson Nano, which sends high-level trajectories to a low-level 
    controller on the target robot platform}
    \label{fig:neuromorphic_stack}
\end{figure}

Figure~\ref{fig:neuromorphic_stack} outlines the proposed \emph{neuromorphic navigation autonomy stack}, which integrates event-driven sensing, real-time mapping, and physics-aware decision-making. The overall architecture is designed to run efficiently on edge hardware (NVIDIA Jetson Nano), ensuring low-latency operation suitable for robotics platforms that require fast responses and minimal energy consumption. 

\subsection{Event-Driven Perception, Localization, and Mapping}

A Dynamic Vision Sensor (DVS) and an Inertial Measurement Unit (IMU) form the core of the event-based sensing layer. Unlike conventional cameras that capture full image frames at fixed intervals, the DVS asynchronously records only changes in pixel brightness at microsecond resolution. This results in a sparse, temporally precise data stream that naturally aligns with spiking neural network (SNN) processing.

The event rate of a moving object is directly proportional to its speed, as rapid brightness changes generate denser event bursts. This leads to faster accumulation of charge in spiking neurons, causing them to fire once the membrane potential surpasses a threshold \(V_{th}\). Figure~\ref{fig:snn_object_detection} illustrates how spiking neurons process incoming events by filtering motion-based features in real time.

To localize objects within the event stream after a neuron fires, a bounding box is extracted based on the spatial distribution of spiking events. This is achieved by computing the minimum and maximum coordinates of the detected spikes:

\begin{equation}
{center}_x = X_{\min} + \left[\frac{X_{\max} - X_{\min}}{2} \right],
\end{equation}
\begin{equation}
{center}_y = Y_{\min} + \left[\frac{Y_{\max} - Y_{\min}}{2} \right].
\end{equation}

These coordinates represent the approximate center of the detected object, providing an efficient method for tracking moving targets in real time. The single-layer SNN is capable of localizing objects even at greater depths, where event density naturally decreases. A weighted neighborhood function with a \(3 \times 3\) kernel size is used for event grouping. The model is fine-tuned using \(\beta = 0.1\) and \(V_{th} = 1.75\), achieving reliable tracking at an object velocity of 4 m/s.

Figure~\ref{fig:ev_box} demonstrates the effectiveness of this method by showing event-based object detection at varying depths. Despite decreasing event density as the object moves further away, the system continues to robustly track and localize the target.

By integrating IMU data for motion estimation and using event-based feature tracking, the system incrementally builds a dynamic map of the environment. For this work, we use an Optitrack Motion Capture \cite{optitrack} system to gather information about the environment while building the initial map. The asynchronous nature of event updates allows it to refine the representation in real time subsequently during local replanning.

\subsection{Physics-Aware Planning and Decision Making}

All high-level navigation logic is executed on the Jetson Nano, allowing the system to combine neuromorphic perception with \emph{physics-informed} planning. By incorporating a lightweight model of the robot’s motion constraints, the planner can generate feasible trajectories that respect velocity, acceleration, and turning limits. This ensures that commands are not only collision-free but also dynamically attainable.

\begin{itemize}
    \item \textbf{Adaptive Planning Algorithms:} The planner can be implemented using classical approaches (e.g., A*, RRT variants) or custom  algorithms \cite{evplanner}. Crucially, updates can occur whenever new perceptual events arrive, permitting rapid re-planning in dynamic environments.
    \item \textbf{Robot Physics Module:} A configurable kinematic or dynamic model (e.g., unicycle, differential drive, or aerial platform) is used to ensure that the motion commands generated by the planner remain valid for the target robot. 
    \item \textbf{High-Level Commands to Low-Level Control:} The planner outputs trajectory or waypoint information at an abstract level, which the robot’s onboard controller uses to drive motors or actuators. 
\end{itemize}

\subsection{Modular and Reconfigurable Architecture}

An important advantage of this framework is its \emph{generic} design: substituting the robot physics module or changing the planning algorithm requires no fundamental changes to the rest of the stack. As long as the perception component provides real-time event data and the planner respects the chosen physics model, the system as a whole remains operational. This is especially beneficial when upgrading hardware or switching from one robot platform to another.

\begin{itemize}
    \item \textbf{Flexible Robot Physics:} Different motion models can be plugged in (e.g., ground rover, multi-rotor drone, manipulator), with the same neuromorphic perception seamlessly feeding into planning.
    \item \textbf{Pluggable Planning Methods:} Swapping out an RRT-based planner for a graph-search method or a learned spiking policy simply changes how trajectories are computed, leaving the perception and control interfaces intact.
    \item \textbf{Edge Deployment on Jetson Nano:} By running all high-level tasks (sensing, mapping, and planning) on a compact, low-power platform, the system is well-suited for mobile deployments and can operate in real time without a cloud connection.
\end{itemize}

The proposed neuromorphic navigation autonomy stack integrates event-based sensing, spiking perception, physics-based planning, and a reconfigurable architecture into one cohesive design. This approach leverages the strengths of neuromorphic sensors (low latency, high dynamic range) while preserving the flexibility to adapt motion models or planning algorithms for new missions and platforms.

\section{Robot-Specific Considerations}
\label{taxonomical}
Robotic platforms differ widely in their degrees of freedom, actuation complexity, and operating domains, necessitating distinct navigation and control strategies. Aerial robots, for instance, typically have more degrees of freedom and tighter inter-actuator coupling than ground-based systems, resulting in increased complexity. Moreover, actuation can range from underactuated to overactuated, influencing agility, fault tolerance, and attainable control authority \cite{valasek2002design}. Physics-informed models and human-intuition priors can guide neural networks to enhance interpretability, robustness, and data efficiency \cite{Raissi2019,rampnet,chee2022knode,shire}. In the following sections, we outline the unique challenges associated with ground and aerial robots and discuss how neuromorphic learning strategies can address them.

\subsection{Ground-based Robots}\label{sec:ground_robots}

\begin{figure}[!t]
    \centering
    \includegraphics[width=0.7\linewidth]{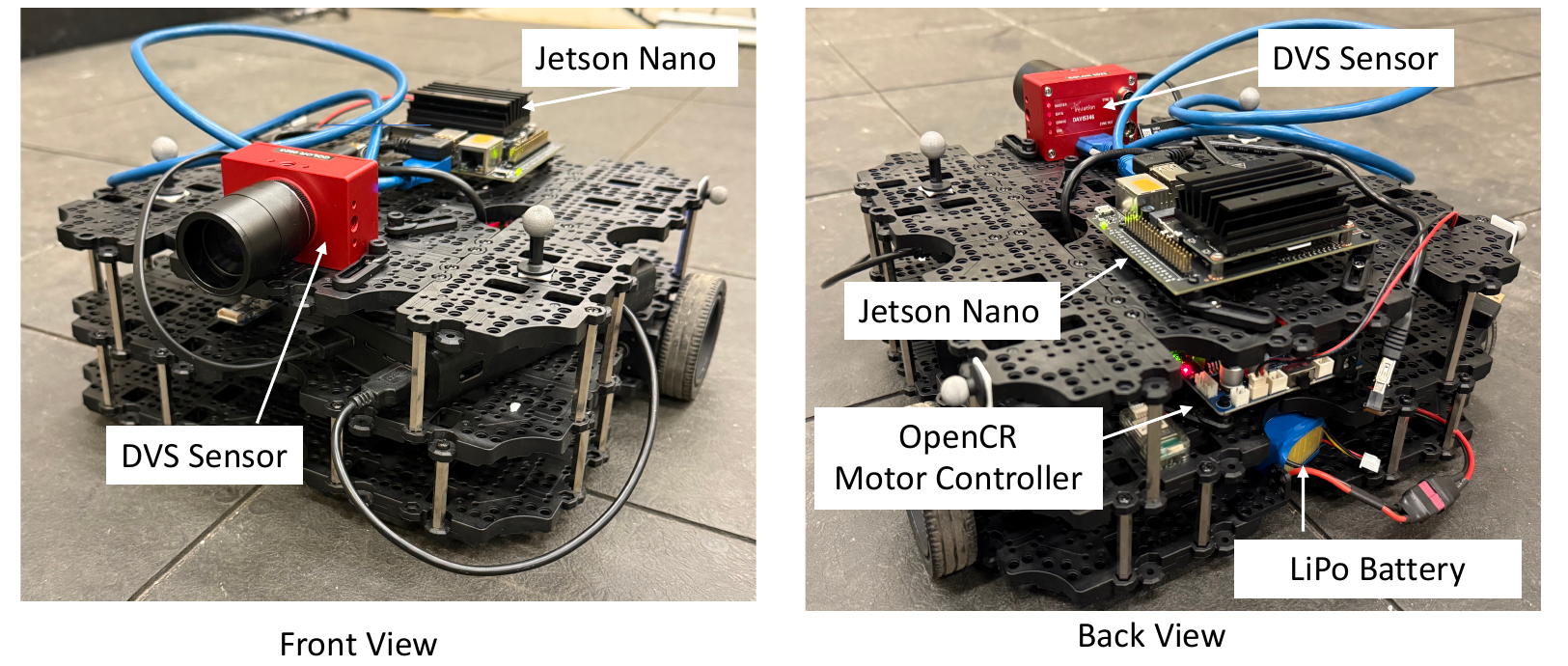}
    \caption{\textbf{Experimental setup for our ground robotics platform}: Front (left) and back (right) views of our TurtleBot-based ground 
    robot platform. A DAVIS346 DVS sensor is mounted on the front for 
    event-driven perception. An NVIDIA Jetson Nano provides onboard 
    neuromorphic computation and high-level planning, while an OpenCR motor 
    controller handles low-level actuation. Power is supplied by a LiPo battery.}
    \label{fig:bot_setup}
\end{figure}

Figure~\ref{fig:bot_setup} shows our demonstration platform, a custom TurtleBot-based system equipped with a DAVIS346 DVS camera for event-driven perception, an NVIDIA Jetson Nano for onboard neuromorphic computing, and an OpenCR board for motor control. This setup illustrates how a ground robot can incorporate neuromorphic sensors and edge computing to react more quickly and efficiently to changes in its environment—particularly when time-critical detection and planning updates are necessary for safe navigation.

Ground-based robots typically have between three and five degrees of freedom, with three being the most common (i.e., two planar translations plus rotation about the vertical axis). This simpler configuration reduces the complexity of the navigation stack, since translational and rotational degrees of freedom can often be treated somewhat independently. However, real-world actuation introduces couplings that complicate idealized control assumptions.

Sensing modalities on ground robots are chosen based on application demands, size, and power constraints. Smaller robots operating in relatively static environments tend to use lightweight sensors (e.g., SONAR, RADAR, IMUs, or indoor localization systems such as UWB and motion capture). Larger robots, particularly those in dynamic domains, augment these sensors with vision (RGB, Depth) to better perceive complex surroundings.

\subsection{Aerial Robots}\label{sec:aerial_robots}

\begin{figure}[!t]
\centering
\includegraphics[width=0.7\linewidth]{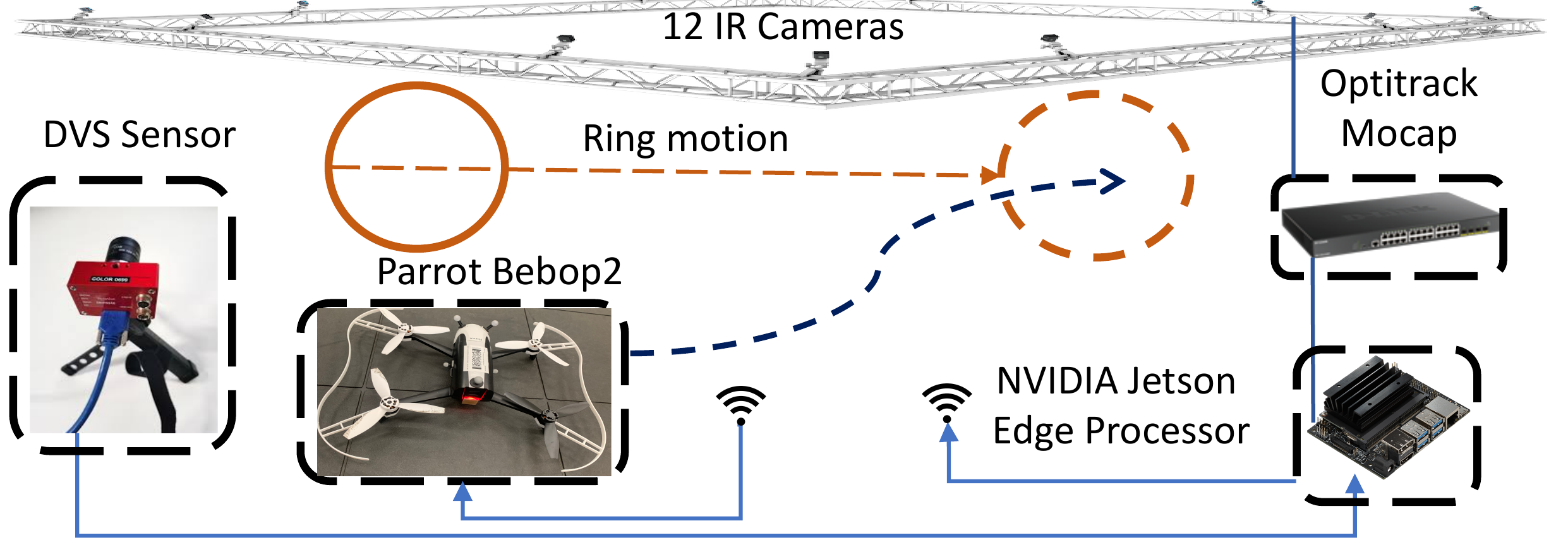}
\caption{\textbf{Experimental setup for our aerial robotics platform}: A Parrot
Bebop2 drone equipped with a DVS camera is tracked by a 12-camera OptiTrack
motion capture system \cite{optitrack}. An NVIDIA Jetson Edge Processor handles event-based
data streams and coordinates real-time ring motion, enabling physics-informed
navigation and control.}
\label{fig:bebop_setup}
\end{figure}

Aerial robots (or drones) possess the ability to translate and rotate about all three axes in three-dimensional space, giving them six degrees of freedom. While this confers agility, it also introduces high cross-coupling among translational and rotational modes and increases the number of possible failure modes. These factors make the design of controllers for aerial robots more complex than for typical ground vehicles. Figure~\ref{fig:bebop_setup} highlights our experimental aerial platform—a Parrot Bebop2 equipped with a DVS camera and tracked by OptiTrack. This setup demonstrates how neuromorphic event-based sensing can be integrated with motion capture data and edge computing (Jetson Nano) to enable real-time, physics-aware planning and control.

Similar to ground-based systems, aerial robots choose sensing modalities based on their mission, size, and power constraints. However, their more intricate dynamics mandate high-resolution IMUs, barometers, rangefinders, and often standard RGB or high-power RADAR sensors. Early commercial aerial robots typically relied on static path planning \cite{lavalle1998rapidly}, waypoint navigation \cite{hrabar2011reactive}, and relatively simple controllers like PID/cascaded PID \cite{hoffmann2007quadrotor, pounds2006modelling}, MPC \cite{kothare1996robust}, or LQR \cite{bouabdallah2004pid}. Over the last decade, sampling-based planners \cite{RRTStar2022}, neural motion planners \cite{penicka2022learning}, and more predictive controllers have gained popularity. Nevertheless, these learning-based methods can fail in out-of-distribution conditions \cite{farid2022task} or fail to incorporate drone \emph{physics} thoroughly \cite{petrik2022learning}.\\

\subsection{Ground Robot Simulation}

Ground robots, particularly differential drive platforms like TurtleBot, follow kinematic constraints that define their motion feasibility. Unlike aerial robots, where dynamics are dominated by force and torque balancing, wheeled robots must adhere to nonholonomic constraints that restrict their motion. The unicycle kinematic model describes this as:
\begin{equation}
\dot{x} = v \cos(\theta), \quad \dot{y} = v \sin(\theta), \quad \dot{\theta} = \frac{v}{L} \tan(\phi),
\end{equation}
where:
\begin{itemize}
    \item \( (x,y) \) represents the robot’s position,
    \item \( v \) is the linear velocity,
    \item \( \theta \) is the heading angle,
    \item \( \phi \) is the steering angle,
    \item \( L \) is the wheelbase length.
\end{itemize}

For energy-aware motion planning, the robot’s power consumption is closely related to rolling resistance, motor efficiency, and dynamic acceleration costs. The actuation energy consumption can be approximated as:
\begin{equation}
P(t) = f_r v + c_a v^3,
\end{equation}
where:
\begin{itemize}
    \item \( f_r \) is the rolling resistance coefficient,
    \item \( c_a \) is the aerodynamic drag coefficient,
    \item \( v \) is the robot’s velocity.
\end{itemize}
Since low-speed ground robots operate in a regime where rolling resistance dominates, energy-efficient navigation strategies often minimize unnecessary velocity fluctuations.

To quantify control effort, we define the torque required to maintain a given acceleration:
\begin{equation}
\tau = I \dot{\omega} + f_d,
\end{equation}
where:
\begin{itemize}
    \item \( I \) is the moment of inertia of the wheels,
    \item \( \dot{\omega} \) is the angular acceleration,
    \item \( f_d \) represents drivetrain losses.
\end{itemize}

\begin{figure}[!t]
    \centering
    \includegraphics[width=0.75\linewidth]{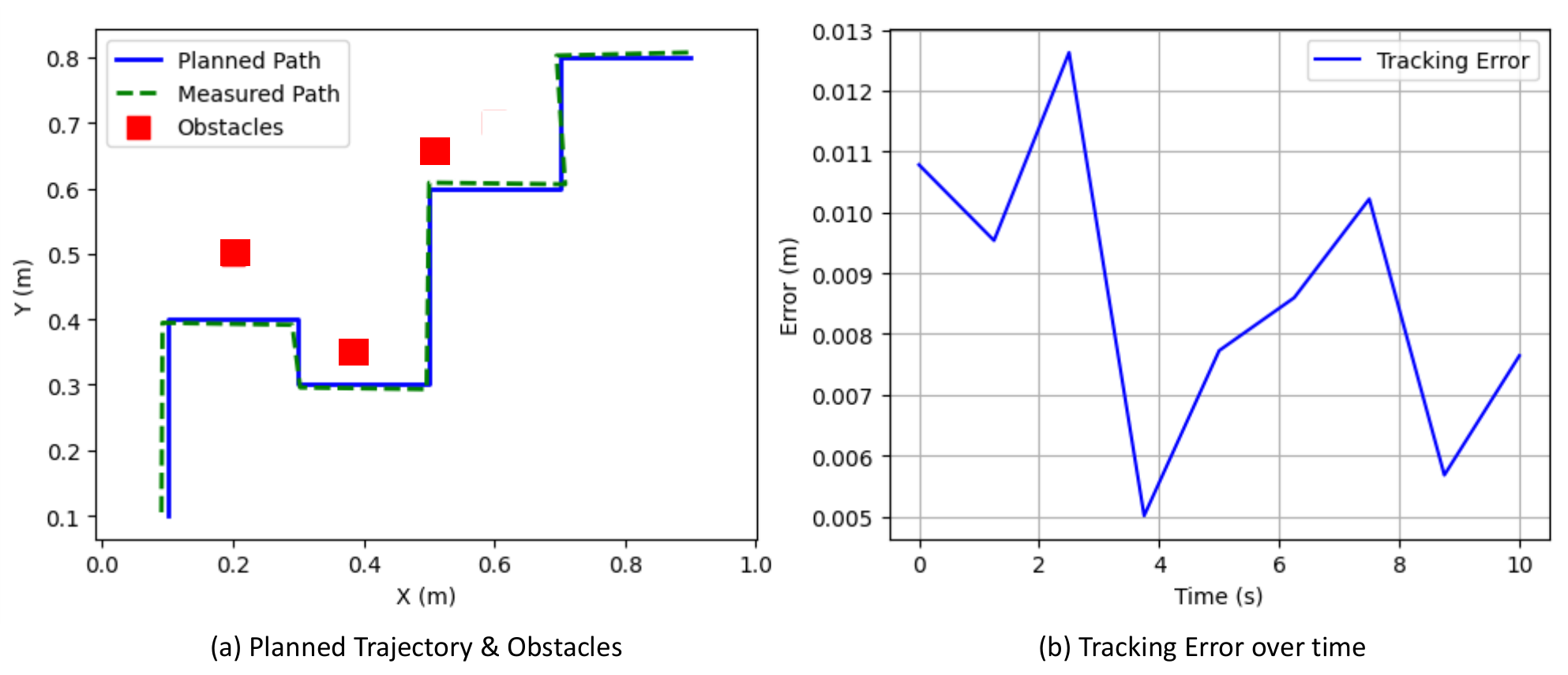} 
    \caption{Example simulation results for our ground robot in a dynamic 
    environment. (a) shows the robot’s planned trajectory (blue) avoiding 
    moving obstacles (red). (b) presents the path tracking error  
    showing reasonable tracking.}
    \label{fig:bot_sim}
\end{figure}

To validate the performance of our neuromorphic navigation on ground-based platforms, we conducted simulation experiments with dynamic obstacles, as shown in Figure~\ref{fig:bot_sim}. In these tests, the TurtleBot model, equipped with simulated event-driven sensors, continuously updated its local map and planned collision-free paths. As shown in Figure~\ref{fig:bot_sim}(b), the event-based strategy exhibited low path tracking error, demonstrating the benefits of asynchronous sensing.

The motion of the TurtleBot is constrained by its differential drive kinematics, where the ability to change direction depends on wheel velocity differences rather than independent lateral control. Unlike aerial robots, which require active thrust modulation to counteract external disturbances, ground robots primarily experience resistive forces such as rolling friction and drivetrain inefficiencies. As a result, the energy consumption of a wheeled robot is dominated by resistive losses and torque demands rather than aerodynamic drag or gravity-compensated thrust forces. Since rolling resistance increases with unnecessary velocity fluctuations, maintaining smooth acceleration and deceleration profiles is critical for energy-efficient motion. Additionally, excessive torque application not only increases motor heating and internal losses but also leads to abrupt trajectory corrections, further impacting actuation energy. These factors emphasize the importance of optimizing control policies to minimize resistive losses while ensuring precise trajectory tracking.

\subsection{Aerial Robot Simulation: }

Unlike ground robots, aerial robots (e.g., quadrotors) operate in a highly dynamic, underactuated environment where forces and energy optimization play a critical role. Many aerial robots have fewer actuators than degrees of freedom, leading to strong nonlinearities and coupling effects \cite{chevallereau2003time, lu2021gain}. Thus, control strategies must ensure feasible maneuvers while minimizing energy consumption.

For battery-limited aerial robots, energy-aware planning is crucial. The power consumption at any instant can be modeled as:
\begin{equation}\label{eq:power_consumption} 
P(t) = \kappa \,\bigl|\mathbf{F}_{{thrust}}(t)\bigr|^\alpha,   
\end{equation}
where:
\begin{itemize}
    \item \( \mathbf{F}_{{thrust}}(t) \) represents the thrust force applied by the motors,
    \item \( \kappa \) is a propulsion constant,
    \item \( \alpha \) is a nonlinearity parameter capturing power-thrust characteristics.
\end{itemize}

\begin{figure}[!t]
    \centering
    \includegraphics[width=0.8\textwidth]{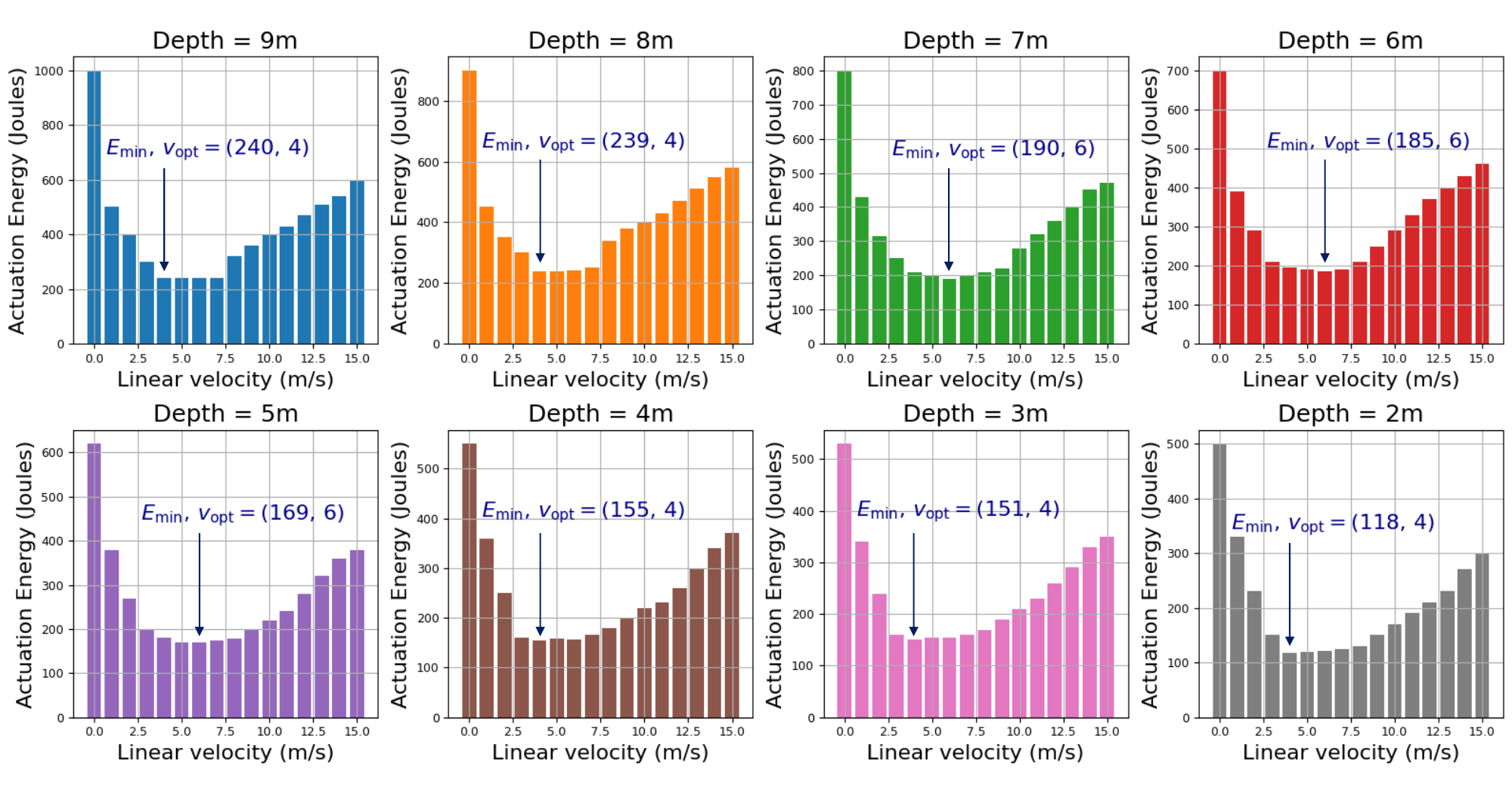}
    \caption{Actuation energy (Joules) vs. linear velocity (m/s) at various depths.}
    \label{fig:energy_time}
\end{figure}

To balance flight time and energy efficiency, the energy-aware objective minimizes an integrated cost:
\begin{equation}\label{eq:energy_objective}
\mathcal{J} = \int_{t_0}^{t_f} \bigl(\beta_1 + \beta_2\,P(t)\bigr)\,\mathrm{d}t,
\end{equation}
where:
\begin{itemize}
    \item \( \beta_1 \) penalizes excessive flight time,
    \item \( \beta_2 \) scales energy consumption.
\end{itemize}

Physics-Informed Neural Networks (PiNNs) and Physics-Guided Neural Networks (PgNNs) integrate these constraints into their loss functions, ensuring learned trajectories are:
\begin{itemize}
    \item Physically feasible by penalizing high-thrust maneuvers.
    \item Energy-efficient by minimizing non-optimal power usage.
    \item Resilient to disturbances by enforcing physics-based stability conditions.
\end{itemize}

Thus, as illustrated in Figure~\ref{fig:energy_time}, motor models help identify optimal velocities (\(v_\mathrm{opt}\)) that minimize energy consumption for different path depths. By incorporating these models as regularizers, neuromorphic planners ensure that real-world trajectories remain both dynamically feasible and energy-efficient.
\begin{figure}[!t]
\centering
\begin{minipage}[t]{\textwidth}
\centering
\includegraphics[width=0.5\textwidth]{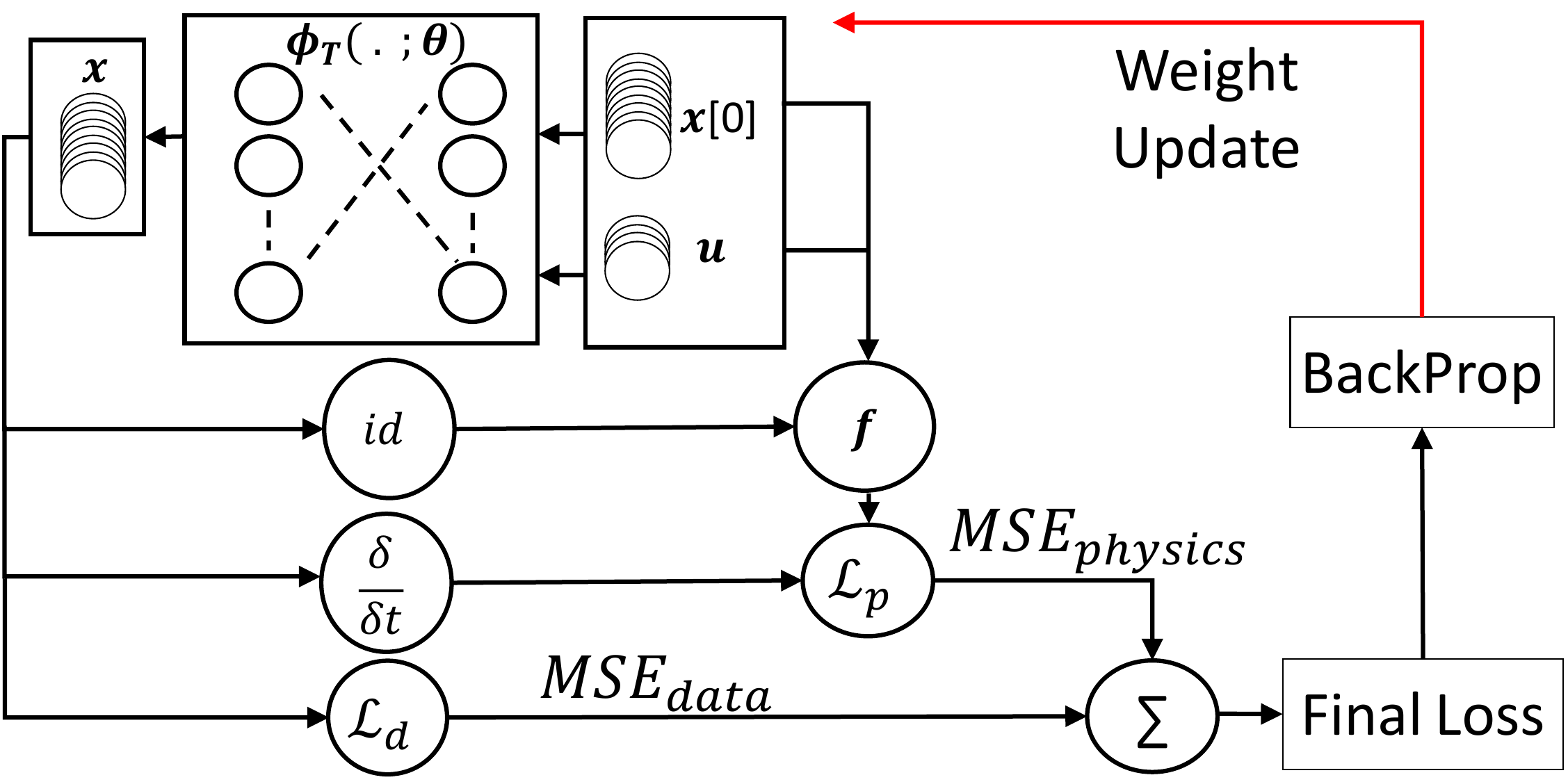}
\end{minipage}
\vspace{0.5cm}

\begin{minipage}[t]{\textwidth}
    \centering
    \includegraphics[width=0.8\textwidth]{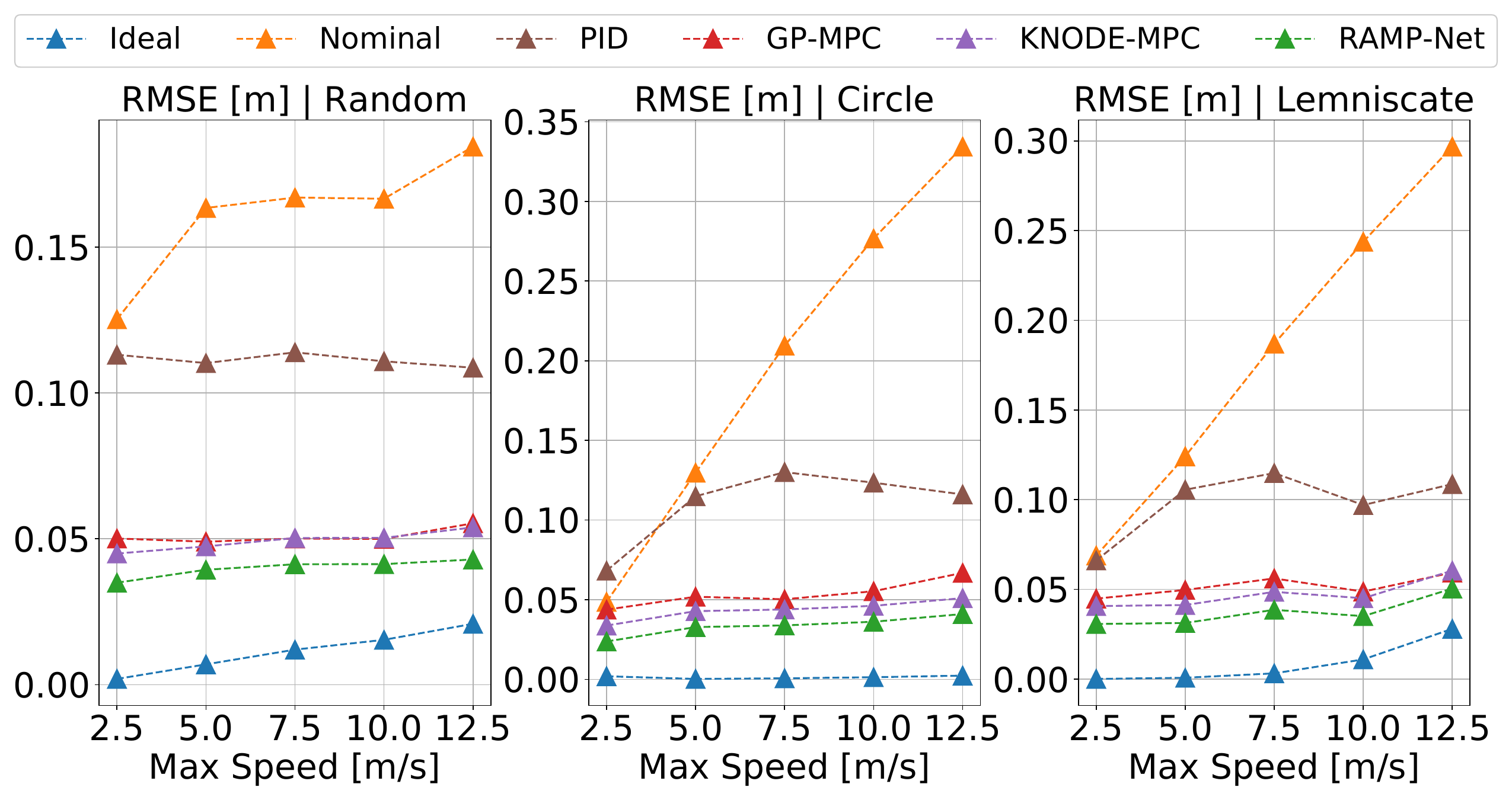}
\end{minipage}
\caption{
    Top: Physics-Informed Neural Networks (PiNNs) embed physical constraints 
    into the loss function, balancing data-based ($\mathcal{L}_\mathrm{data}$) 
    and physics-based ($\mathcal{L}_\mathrm{physics}$) terms. \\
    Bottom: RAMP-Net \cite{rampnet} uses PiNNs for trajectory tracking under 
    dynamic disturbances; RMSE plots compare RAMP-Net to alternative controllers 
    for three trajectory patterns (\textit{Random}, \textit{Circle}, 
    \textit{Lemniscate}), showing superior accuracy.
}
\label{fig:pinn_and_mse}
\end{figure}
Physics-Informed Neural Networks (PiNNs) introduce domain-specific knowledge into learning-based controllers by embedding physical constraints directly into the optimization process. Figure~\ref{fig:pinn_and_mse} (top) illustrates how PiNNs integrate physics-based losses ($\mathcal{L}_\mathrm{physics}$) with conventional data-driven loss terms ($\mathcal{L}_\mathrm{data}$). This approach ensures that the learned policies remain physically consistent, improving robustness in real-world deployment.

The effectiveness of PiNNs is demonstrated in Figure~\ref{fig:pinn_and_mse} (bottom), where RAMP-Net \cite{rampnet} is compared against alternative control strategies for trajectory tracking in dynamic conditions. The RMSE plots illustrate the tracking performance across three trajectory patterns: \textit{Random}, \textit{Circle}, and \textit{Lemniscate}, as a function of the robot's maximum speed. The results show that:
\begin{itemize}
    \item Traditional controllers, such as PID and Nominal control, exhibit increasing error at higher speeds due to their limited adaptability.
    \item Gaussian Process Model Predictive Control (GP-MPC) improves stability but still suffers from tracking degradation at higher speeds.
    \item Physics-Informed Neural Networks (PiNNs), as implemented in RAMP-Net, achieve significantly lower RMSE across all trajectories, demonstrating their ability to maintain precise trajectory tracking even at high speeds.
\end{itemize}

By explicitly enforcing physical constraints within the learning process, PiNN-based controllers outperform conventional methods, particularly in high-speed and dynamically changing environments.

A systematic integration of neuromorphic sensing, physics-informed methods, and energy-aware objectives can potentially elevates the agility and robustness of aerial navigation while significantly reducing computational and power overhead.

\subsection{Real World Demonstrations}
\subsubsection{Ground Robot (TurtleBot): }

To evaluate the real-world performance of our neuromorphic navigation framework, we deployed the TurtleBot in a dynamic indoor environment featuring multiple static obstacles. The robot was equipped with a DAVIS346 event-based vision sensor and relied on an asynchronous spiking neural network for real-time perception. For depth estimation, we employed the divergence method, which infers depth from the spatial and temporal distribution of events. The rate of change in event density provides a reliable cue for relative depth, allowing the system to estimate the distance to surrounding obstacles. This event-based depth inference ensures continuous updates without relying on dense depth maps, making it particularly suited for fast-moving environments.
To separate ego-motion from object motion, we used contrast maximization \cite{gallego2018unifying}, a common method in event-based visual odometry. By maximizing the alignment of consecutive event frames, the system isolates motion caused by the robot’s movement. This approach improves localization accuracy and ensures the robot maintains a consistent understanding of the scene.

\begin{figure}[!t] \centering \includegraphics[width=0.9\linewidth]{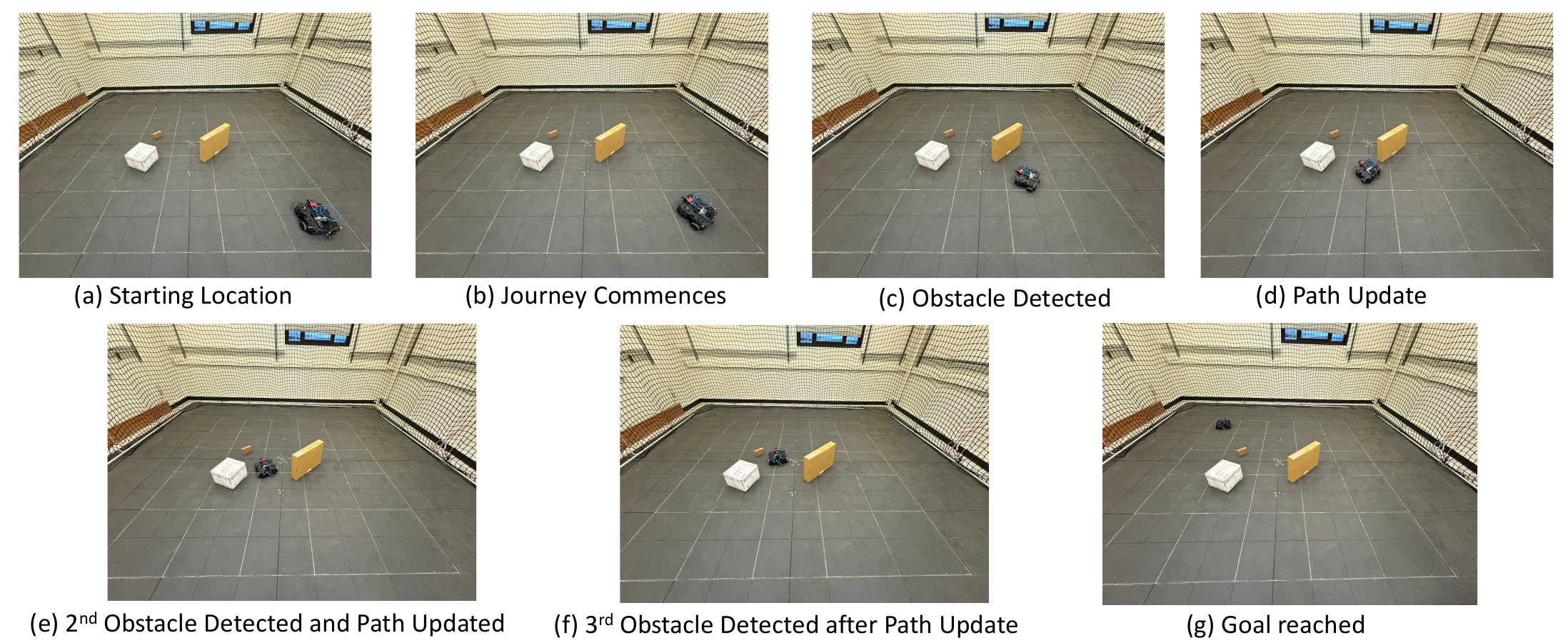} \caption{Real-world TurtleBot navigation using neuromorphic vision. The robot successfully detects and avoids obstacles while dynamically updating its trajectory.} \label{fig:bot_real_traj} \end{figure}

As illustrated in Figure~\ref{fig:bot_real_traj}, the TurtleBot successfully navigated from the start position to the goal while dynamically avoiding obstacles. The sequence of images captures key moments during navigation: the robot first identifies an obstacle (Figure~\ref{fig:bot_real_traj}c), updates its path (Figure~\ref{fig:bot_real_traj}d), detects additional obstacles (Figures~\ref{fig:bot_real_traj}e and \ref{fig:bot_real_traj}f), and ultimately reaches the goal (Figure~\ref{fig:bot_real_traj}g).
The real-world experiment with the TurtleBot demonstrates how neuromorphic navigation enables efficient obstacle avoidance and path replanning. Unlike conventional frame-based methods that suffer from latency in dynamic settings, our event-driven pipeline ensures continuous real-time adjustments.
\begin{figure}[!t]
    \centering
    \includegraphics[width=0.9\linewidth]{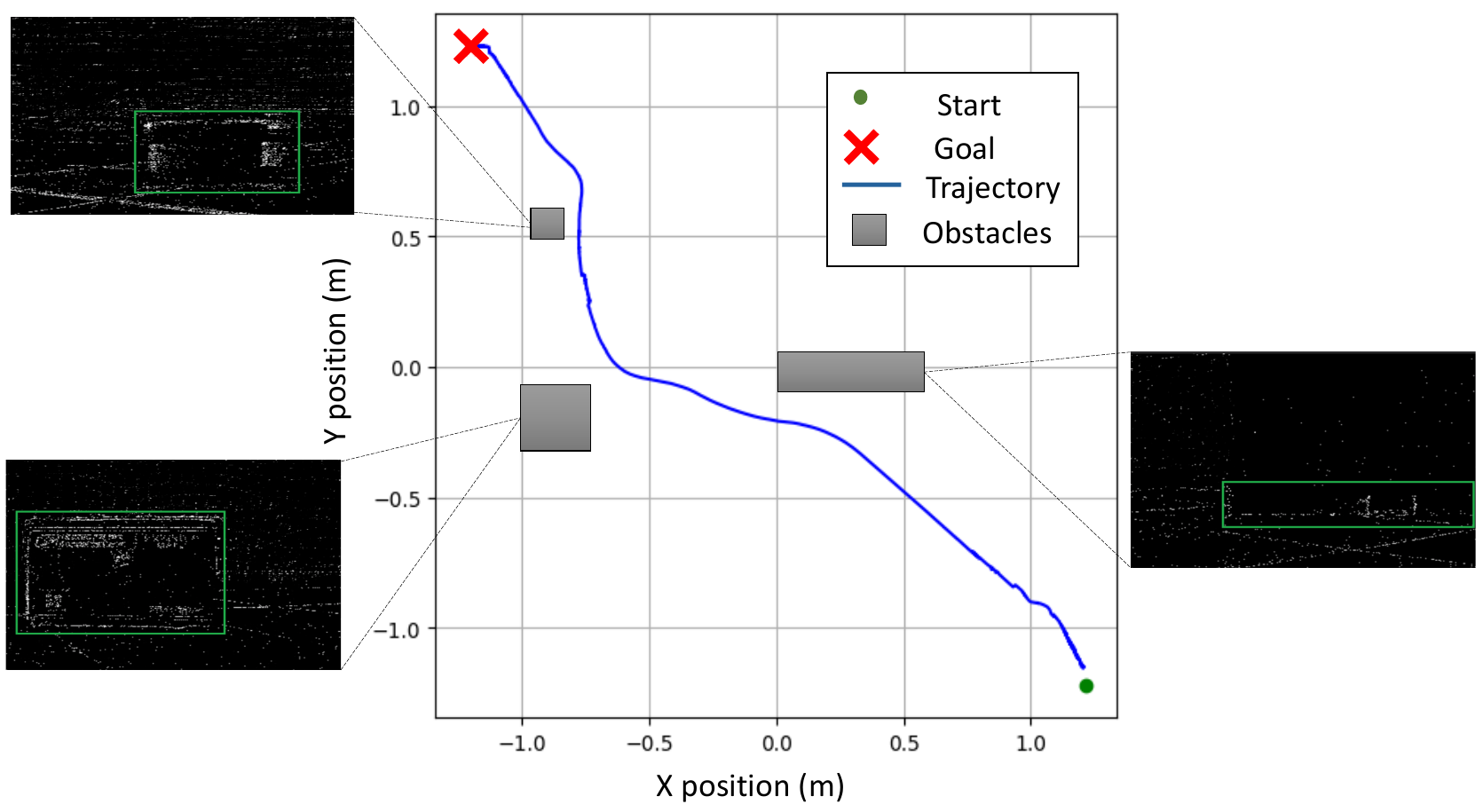} 
    \caption{TurtleBot's real-world trajectory navigating around obstacles using event-based sensing. 
    The robot starts at the green marker and successfully reaches the goal (red cross) while dynamically avoiding obstacles (gray blocks). 
    The blue trajectory demonstrates the planned path with real-time corrections based on neuromorphic vision. The neuromorphic detection renderings are shown for each obstacle in magnified view.}
    \label{fig:bot_plot}
\end{figure}

As seen in Figure~\ref{fig:bot_plot}, the TurtleBot starts from an initial position and dynamically updates its path as it encounters multiple obstacles. The trajectory reveals smooth corrections around obstructions, demonstrating the system’s responsiveness. The use of spiking neural networks (SNNs) and event-based perception significantly enhances the agility of the robot, allowing it to navigate cluttered environments with minimal computational overhead.
These validate the usefulness of neuromorphic sensing for fast, low-power, and adaptive navigation, making it a compelling approach for real-world deployment in robotics applications.

\subsubsection{Aerial Robot (Bebop2): }
To validate the neuromorphic navigation framework in real-world aerial scenarios, we deployed the Parrot Bebop2 quadrotor equipped with a DAVIS346 event-based camera. The quadrotor was tasked with navigating through a dynamic environment featuring a suspended ring, requiring precise detection, tracking, and agile maneuvers. The goal was to assess the ability of event-driven sensing and neuromorphic processing to enhance real-time perception and control, particularly under fast-changing conditions.

The experiment was conducted in a motion-capture-equipped flight arena, where the drone executed a sequence of navigation tasks: initialization, ring tracking, trajectory correction, and landing. Unlike frame-based methods, which process redundant visual data, our event-driven approach allowed the drone to react asynchronously to relevant scene changes, reducing latency and improving agility.

\paragraph{Flight Phases:}
\begin{itemize}
    \item \textbf{Initialization:} The quadrotor performs system checks, stabilizes using onboard IMU, and enters AltHold mode.
    \item \textbf{Ring Detection and Tracking:} The event-based sensor detects the motion of the target ring, continuously updating its position estimate.
    \item \textbf{Maneuvering Through the Ring:} The drone generates an optimized trajectory using physics-informed planning to successfully pass through the ring.
    \item \textbf{Descent and Landing:} The system transitions to a controlled descent phase, ensuring a precise touchdown.
\end{itemize}

\begin{figure*}[!t]
    \centering
    \begin{subfigure}{0.32\textwidth}
        \includegraphics[width=\textwidth]{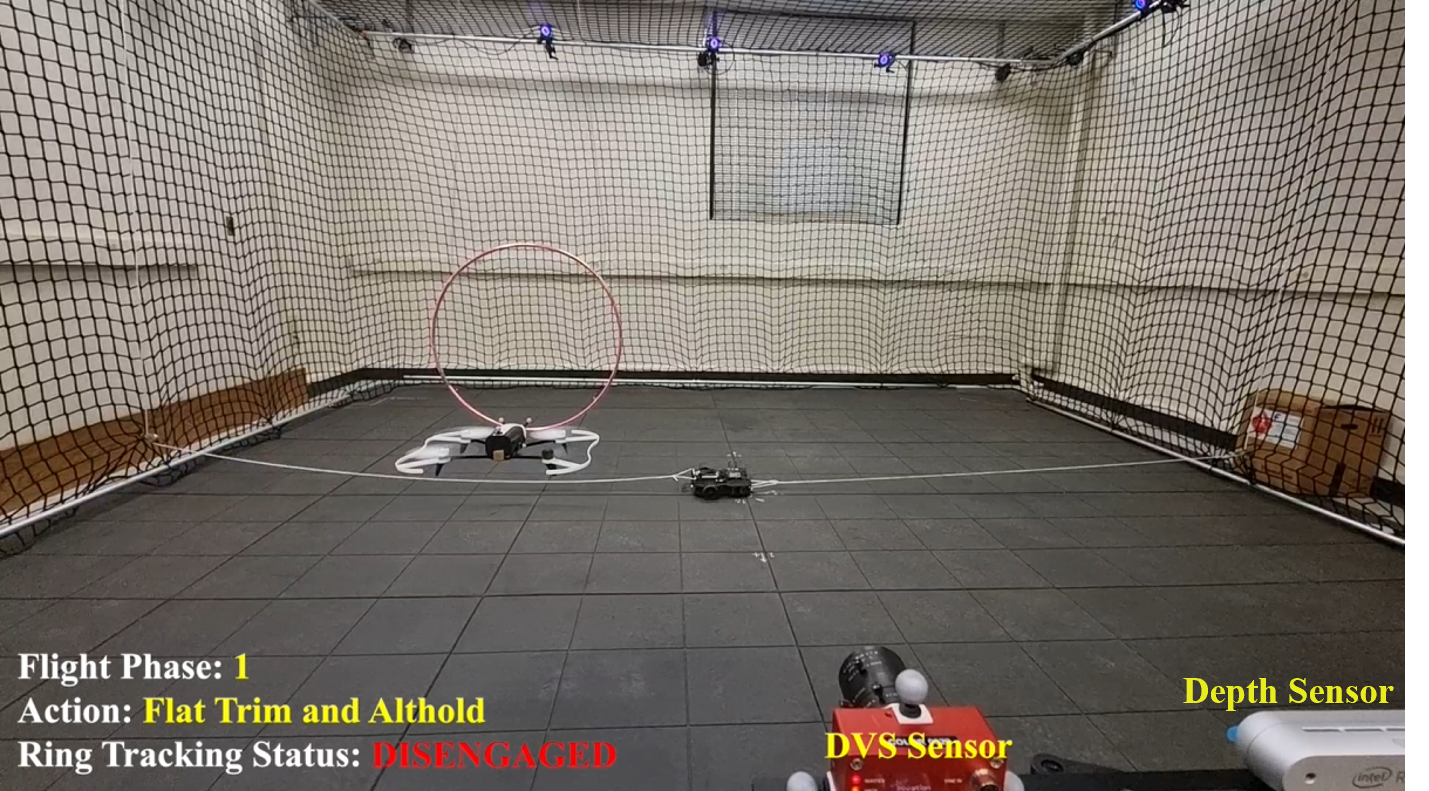}
        \caption{Flat Trim and Althold}
        \label{fig:st1}
    \end{subfigure}
    \hfill
    \begin{subfigure}{0.32\textwidth}
        \includegraphics[width=\textwidth]{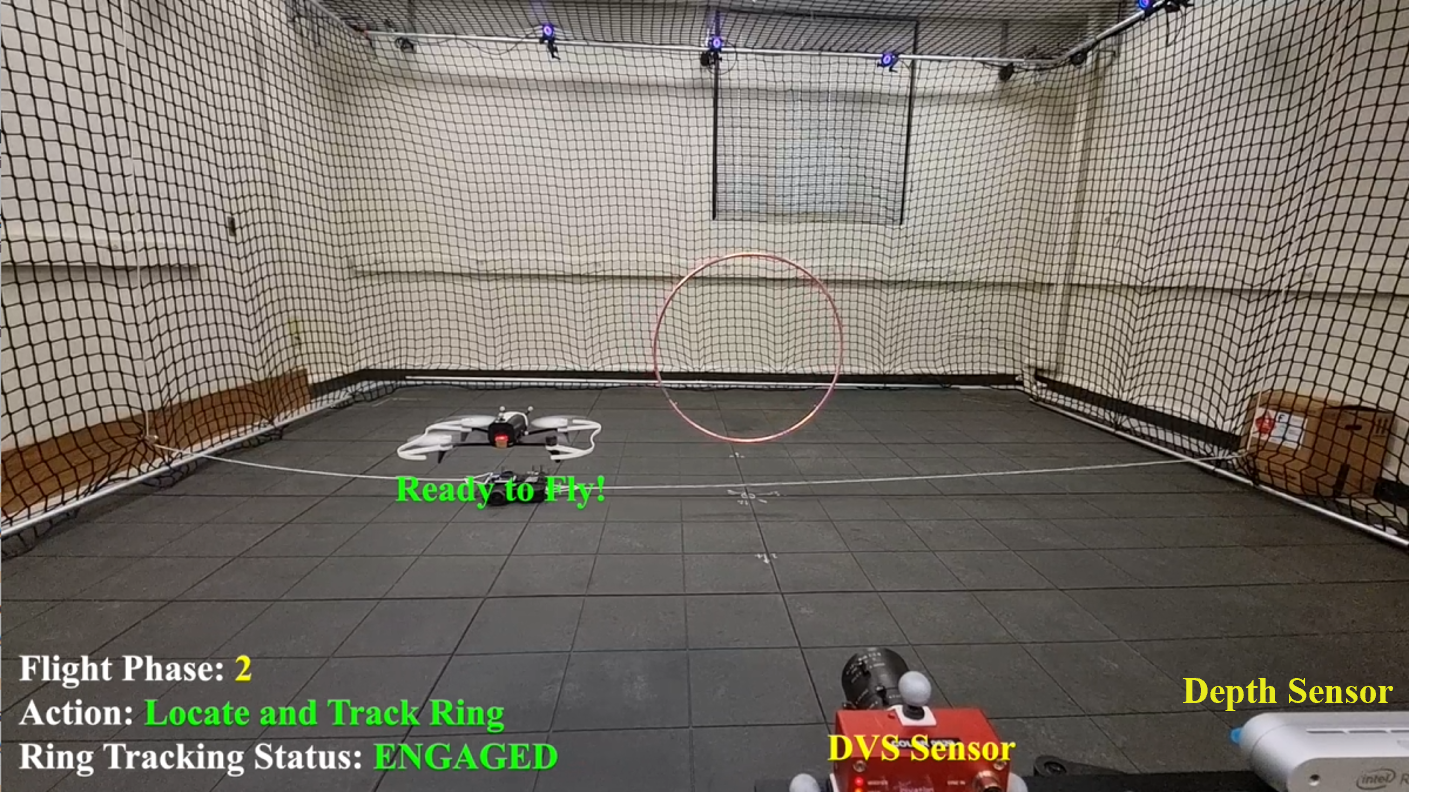}
        \caption{Locate and Track Ring}
        \label{fig:st2}
    \end{subfigure}
    \hfill
    \begin{subfigure}{0.32\textwidth}
        \includegraphics[width=\textwidth]{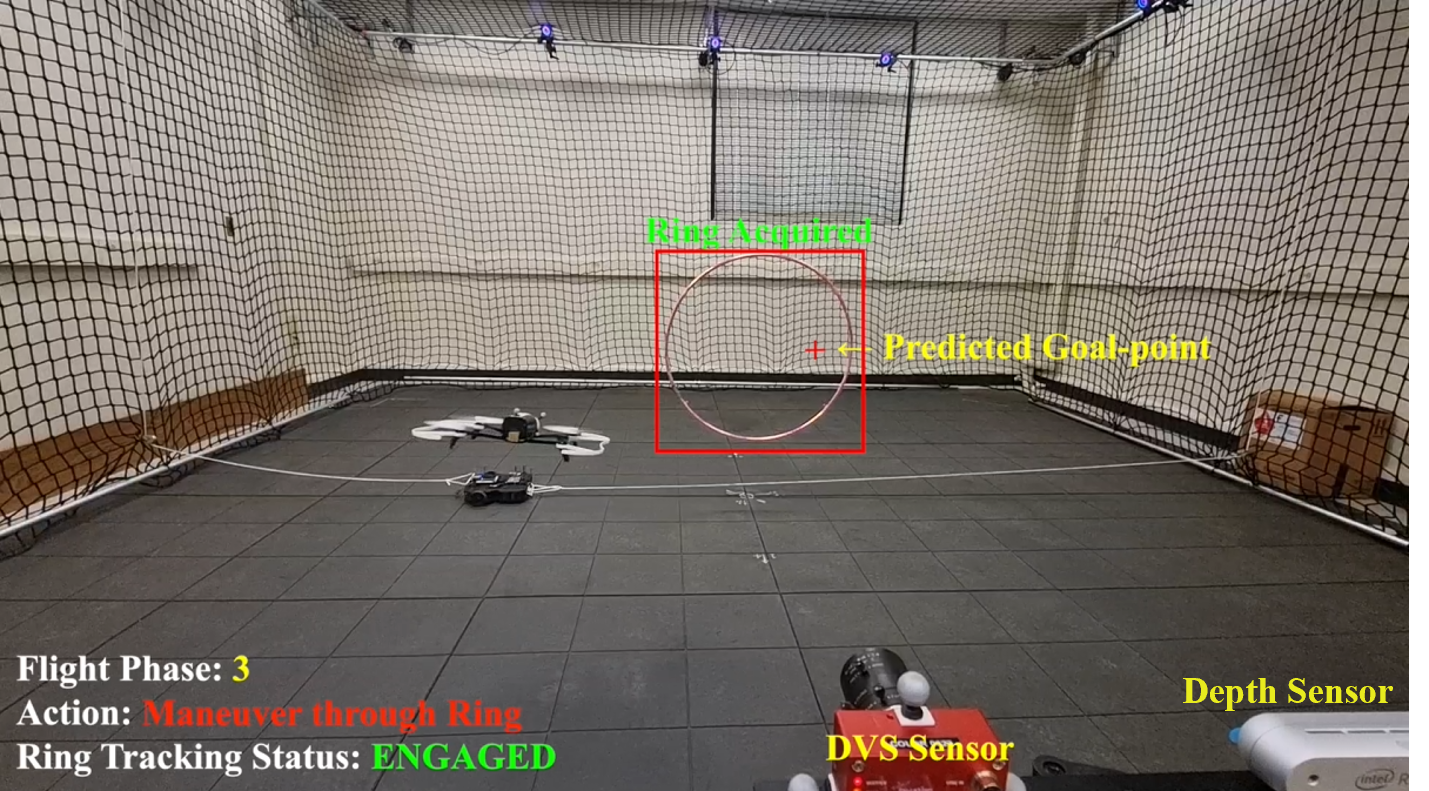}
        \caption{Maneuver through Ring}
        \label{fig:st3}
    \end{subfigure}
    \hspace{177.80093mm}
    \begin{subfigure}[t]{0.32\textwidth}
        \includegraphics[width=\textwidth]{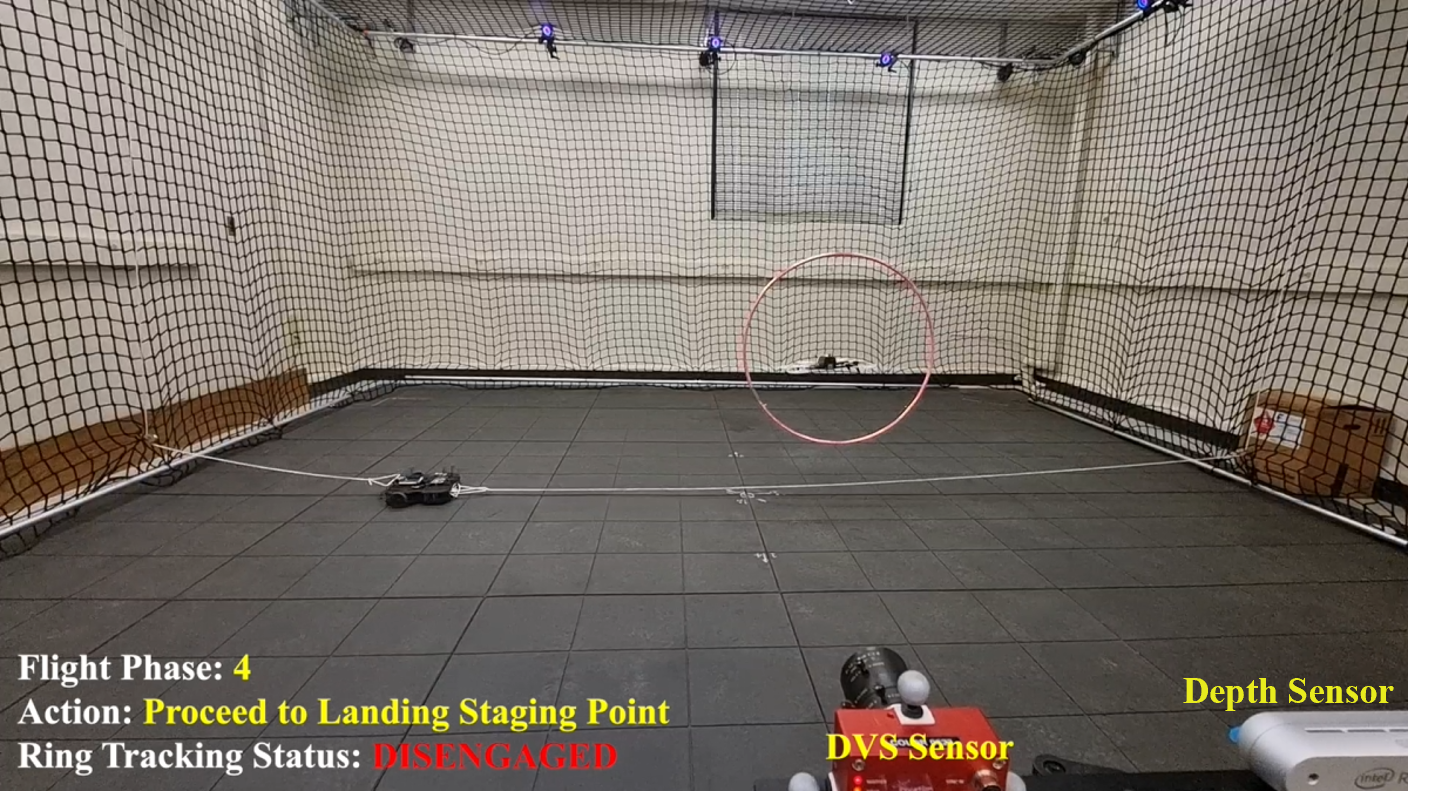}
        \caption{Prepare to Land}
        \label{fig:st4}
    \end{subfigure}
    \hspace{1mm}
    \begin{subfigure}[t]{0.32\textwidth}
        \includegraphics[width=\textwidth]{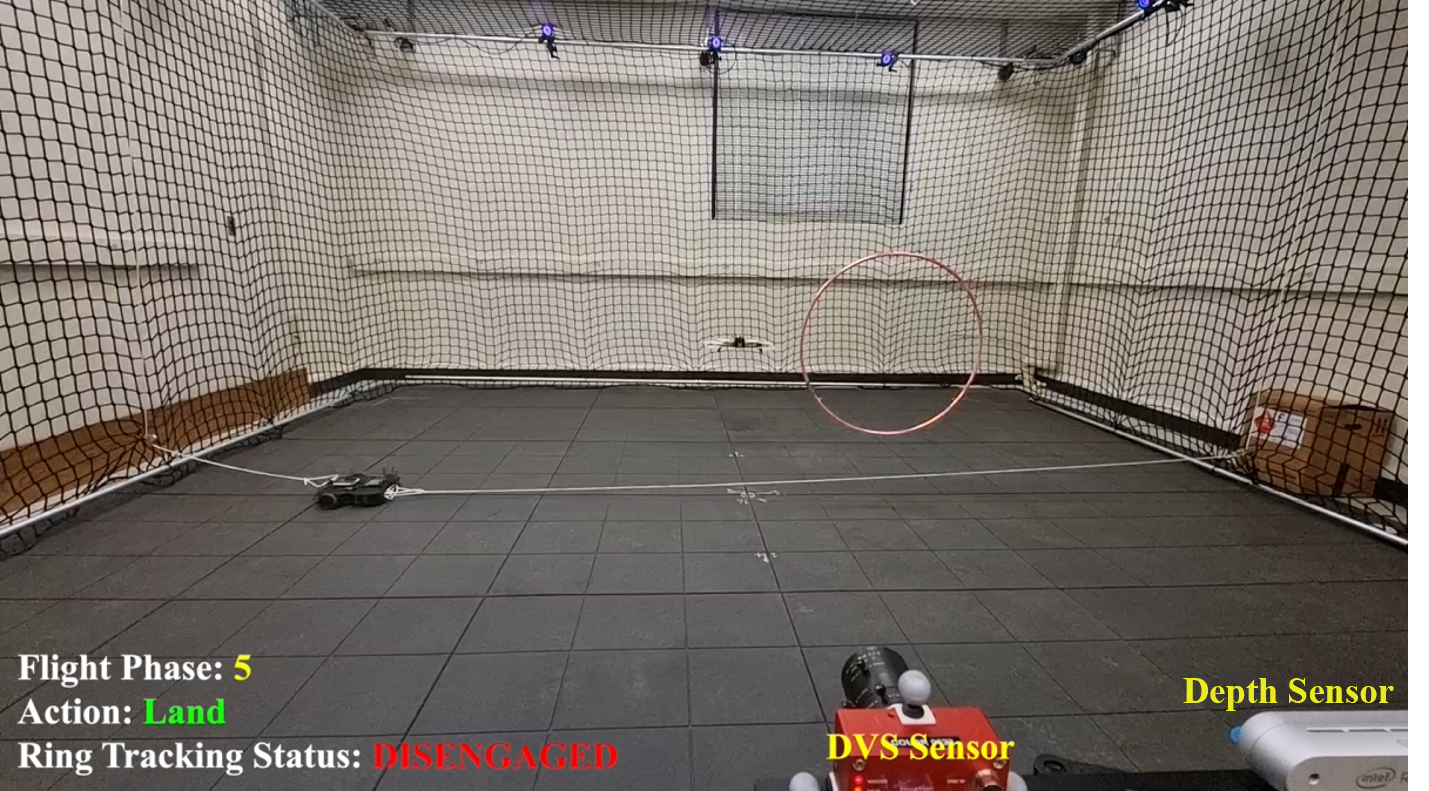}
        \caption{Phase 5: Land}
        \label{fig:st5}
    \end{subfigure}
    \caption{Real-world demonstration of the Bebop2 quadrotor performing neuromorphic navigation through a dynamic environment. 
(a) The drone initializes with flat trim and enters AltHold mode. 
(b) It detects and tracks the target ring using event-based vision. 
(c) The drone autonomously maneuvers through the ring, adjusting its trajectory in real time. 
(d) The landing sequence begins with a controlled descent. 
(e) The final landing phase is executed with precise position control.}

    \label{fig:bebop_real_traj}
\end{figure*}

Figure~\ref{fig:bebop_real_traj} illustrates the different stages of the experiment, highlighting the quadrotor's ability to adapt to dynamic environmental changes. The event-driven sensing pipeline enabled efficient trajectory updates, leading to reasonably visible decision-making.

\begin{figure}[!t]
    \centering
    \includegraphics[width=0.5\linewidth]{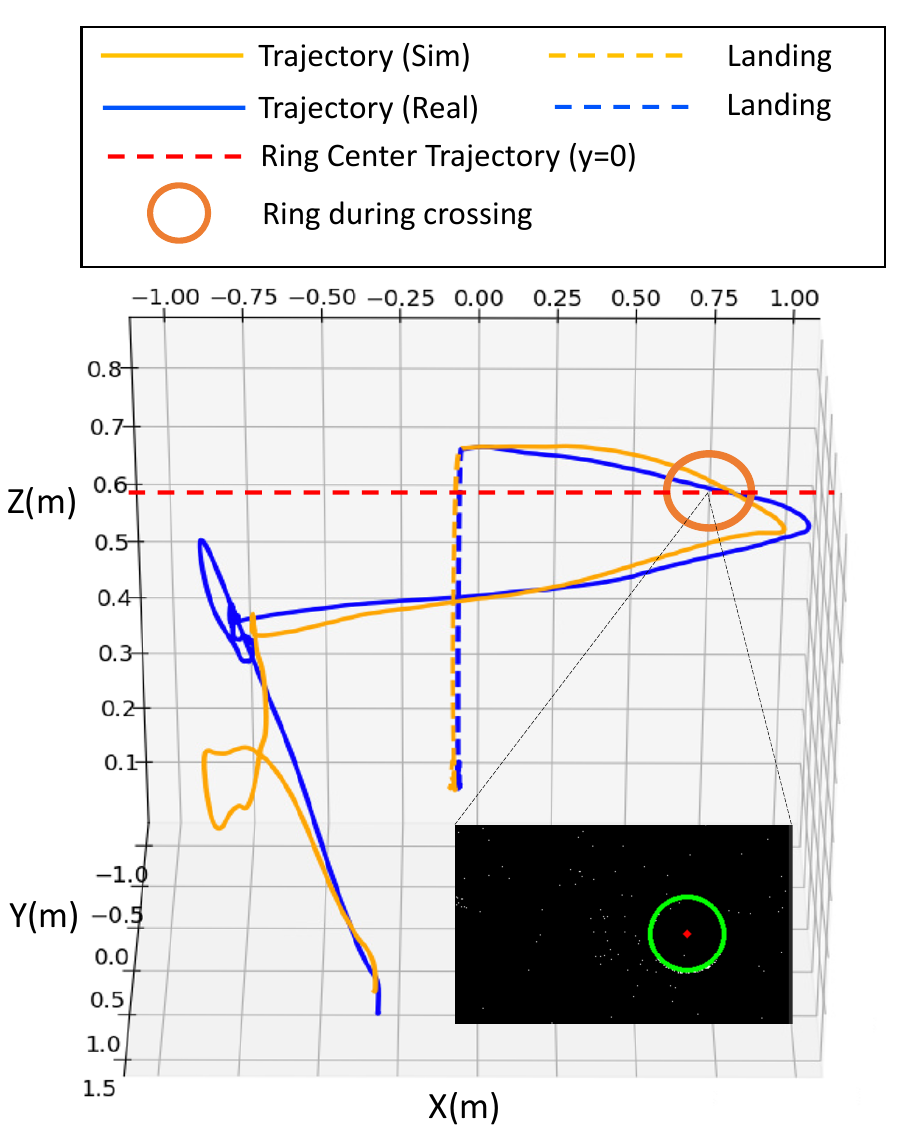}
    \caption{3D trajectory of the Bebop2 drone navigating towards and through the moving ring. The trajectories of the drone for simulated (yellow) and real-world (blue) implementations are shown, with the respective landing phases indicated by dashed lines. The red dashed line represents the ring's center trajectory at a fixed \( y = 0 \) for visualization. The orange circle marks the ring's position at the crossing moment. The neuromorphic detection rendering is shown in the magnified view. 
}
    \label{fig:bebop_traj}
\end{figure}

Figure~\ref{fig:bebop_traj} illustrates the real-world flight trajectory of the Bebop2 drone as it navigates towards and through a moving ring using neuromorphic vision and physics-informed control. The blue trajectory represents the drone’s path, demonstrating its ability to dynamically adjust its flight to align with the moving target. The red dashed linemarks the ring’s center trajectory (fixed at \(y=0\) for visualization), showcasing the planned reference path. At the crossing moment, the orange circle highlights the ring's position relative to the drone.  The flight path traced by real-world implementation deviates from the path traced by simulation primarily due to the inherent non-idealities in motor control and physical constraints not present in the simulation. We also observed the flight and motion paths for the real-world  to be sharper and more abrupt. We present the actuation energies of the two methods using the energy model of the brushless DC motor reported in \cite{evplanner}.
The energy consumption of the real-world flight ($\sim$613 Joules) is found to be $\sim15\%$ higher than its simulated counterpart ($\sim532$ Joules). This increase in energy highlights the challenges and inefficiencies that emerge when transitioning from simulation to actual physical environments. Please note that the energies reported are actuation energies and not the significantly lower perception compute energies.

From the trajectory, we observe that the drone successfully tracks and intercepts the ring's movement, dynamically adjusting its altitude and approach angle. The ability to react to moving obstacles in real time is crucial for aerial robots operating in dynamic environments, such as agile navigation through cluttered spaces or autonomous drone racing. The deviations in the trajectory also reflect the impact of real-world disturbances, sensor latency, and aerodynamic effects, further reinforcing the importance of robust neuromorphic control.

\subsection{Latency, Energy Efficiency, and Success Rate Comparison}
To comprehensively evaluate the proposed neuromorphic navigation framework, we compare its performance against conventional RGB-based navigation and intermediate neuromorphic configurations. Table~\ref{tab:performance_comparison} provides a quantitative comparison of maneuver time, actuation energy, and compute energy across different configurations for both the Bebop2 drone and TurtleBot.

\begin{table*}[t]
    \centering
    \caption{Performance Comparison Across Different Navigation Configurations}
    \label{tab:performance_comparison}
    \renewcommand{\arraystretch}{1.2}
    \setlength{\tabcolsep}{6pt} 
    \scriptsize 
    \begin{tabular*}{\textwidth}{@{\extracolsep{\fill}} l p{1.5cm} ccc @{}}
        \toprule
        \textbf{Configuration} & \textbf{Platform} & \multicolumn{1}{c}{\textbf{Maneuver Time}} & \multicolumn{1}{c}{\textbf{Actuation Energy}} & \multicolumn{1}{c}{\textbf{Compute Energy}} \\
        & & \multicolumn{1}{c}{\textbf{(s) $\downarrow$}} & \multicolumn{1}{c}{\textbf{(J) $\downarrow$}} & \multicolumn{1}{c}{\textbf{(mJ) $\downarrow$}} \\
        \midrule
        RGB-Based Navigation & Bebop2 & 35.3 $\pm$ 5.2 & 845 $\pm$ 35 & 93.5 $\pm$ 4.5 \\
        Neuromorphic Perception Only & Bebop2 & 32.9 $\pm$ 3.5 & 705 $\pm$ 22 & 74.2 $\pm$ 3.8 \\
        Full Neuromorphic Stack & Bebop2 & \textbf{25.3} $\pm$ \textbf{2.0} & \textbf{613} $\pm$ \textbf{14} & \textbf{69.2} $\pm$ \textbf{3.5} \\
        \midrule
        RGB-Based Navigation & TurtleBot & 63.4 $\pm$ 4.7 & 1710 $\pm$ 28 & 93.6 $\pm$ 4.3 \\
        Neuromorphic Perception Only & TurtleBot & 64.7 $\pm$ 3.8 & 1602 $\pm$ 30 & 88.7 $\pm$ 4.2 \\
        Full Neuromorphic Stack & TurtleBot & \textbf{49.8} $\pm$ \textbf{2.0} & \textbf{1373} $\pm$ \textbf{12} & \textbf{74.5} $\pm$ \textbf{3.7} \\
        \bottomrule
    \end{tabular*}
\end{table*}

The results clearly demonstrate that the full neuromorphic stack, which integrates perception, planning, and control, significantly outperforms RGB-based navigation. This is evident in the lower maneuver time and improved energy efficiency, emphasizing the advantages of event-based vision and neuromorphic processing. While neuromorphic perception alone yields a notable reduction in maneuver time and energy consumption, the greatest efficiency gains occur when neuromorphic perception is combined with physics-informed planning and adaptive neuromorphic control. For RGB-based Navigation, we used the YOLO framework \cite{diwan2023object} framework which consists of 60-65 million parameters.
As summarized in Table~\ref{tab:comparison}, our neuromorphic detection approach has an extremely low parameter count (one \(3\times 3\) convolutional layer and 9 spiking neurons). It operates without a large labeled dataset but still requires threshold tuning (e.g., spiking neuron firing thresholds). Our SNN maintains a balance of minimal parameter overhead, unsupervised learning, while offering the benefits of neuromorphic perception using SNN-based dynamic vision sensors.

Although our neuromorphic perception framework uses significantly fewer parameters compared to YOLO, the reduction in compute energy is not as pronounced as the difference in model size. This is primarily due to the reliance on standard CUDA kernels on the NVIDIA Jetson platform, which introduces a considerable setup time during kernel initialization. To achieve even lower compute energy than currently reported, there is a clear need for dedicated neuromorphic hardware. Such hardware would enable greater energy efficiency by minimizing overhead and optimizing event-driven processing at the hardware level, shown by \cite{sharma2024spidr, trackingsoclele, vidal2018ultimate, toffe}.
\begin{table*}[tb]
\centering
\caption{Comparison of tracking methods }
\label{tab:comparison}
\resizebox{1.0\linewidth}{!}{
\begin{tabular}{lcccc}
\toprule
\textbf{Method} & 
\textbf{Parameter Count} & 
\textbf{Training / Labeling} & 
\textbf{Thresholds} & 
\textbf{Notes} \\
\midrule
\textbf{SNN \cite{nagaraj2023dotie,evplanner,joshi2024}} & 
\begin{tabular}[c]{@{}l@{}}
1 Conv2D \((3\times3)\)\\
9 Spiking Neurons
\end{tabular} & 
Unsupervised & 
\begin{tabular}[c]{@{}l@{}}
Yes (firing \\
threshold)
\end{tabular} &
Lightweight, event-driven \\
\midrule
\textbf{YOLO (CNN-based) \cite{diwan2023object}} & 
\(\sim\!60{-}65\)M params & 
\begin{tabular}[c]{@{}l@{}}
Requires large \\
labeled dataset
\end{tabular} &
No &
Popular real-time detector \\
\bottomrule
\end{tabular}%
}
\end{table*}

Furthermore, the results suggest a direct correlation between perception latency, maneuver time, and energy efficiency. Since neuromorphic vision enables faster perception, it leads to a reduction in maneuver time, thereby shortening the overall path length. This, in turn, results in lower actuation energy, as less power is required to complete the navigation task. This cascade effect underscores the fundamental advantage of event-driven neuromorphic perception in reducing both computational and physical energy costs, making it a compelling choice for low-power robotic applications.

\section{Conclusion}
\label{conclusion}
We presented a neuromorphic navigation framework that integrates event-based vision with a task-specific, reconfigurable autonomy stack for real-time robotic navigation. By leveraging neuromorphic sensing, spiking neural networks, and physics-informed planning, our approach enables efficient perception, adaptive decision-making, and energy-aware control across diverse robotic platforms. Through both simulation-based validation and real-world demonstrations, we showed the effectiveness of our framework in navigating dynamic environments with both ground (TurtleBot) and aerial (Bebop2) robots. Our results highlight the advantages of event-driven perception in reducing latency and computational overhead while improving responsiveness to environmental changes. Additionally, the integration of physics-informed neural networks enhances trajectory optimization, ensuring feasible and energy-efficient motion planning. While this work demonstrates practical feasibility using the Jetson Edge processor for real-time neuromorphic processing, the full potential of neuromorphic algorithms will emerge with next-generation hardware \cite{sharma2024spidr} optimized for ultra-low power and efficient spiking computation. Future advancements in neuromorphic computing are expected to enhance scalability and deployment in real-world robotics, and as large language models (LLMs) advance, integrating neuromorphic navigation systems with LLM-based reasoning frameworks \cite{joshi2025neuro} could enable interactive, human-in-the-loop decision-making for complex robotic tasks
Moving forward, expanding the framework to multi-robot systems, incorporating reinforcement learning for adaptive planning, and optimizing neuromorphic hardware implementations are promising avenues for future work. As neuromorphic technologies continue to evolve, their application in autonomous navigation holds significant potential for developing agile, low-power, and highly responsive robotic systems capable of operating in complex, real-world scenarios.

\printbibliography

\end{document}